\pdfoutput=1
\documentclass[11pt]{article}

\usepackage{ACL2023}

\usepackage{times}
\usepackage{latexsym}
\usepackage{color}
\usepackage{graphicx}
\usepackage{algorithm}
\usepackage{algpseudocode}  
\usepackage{amsmath}

\usepackage{amssymb}
\usepackage{booktabs}
\usepackage{makecell}
\usepackage{multirow}
\usepackage{float}

\usepackage{enumerate}
\usepackage{enumitem}
\usepackage{svg}
\usepackage{colortbl}

\usepackage[T1]{fontenc}
\usepackage[utf8]{inputenc}

\usepackage{microtype}

\usepackage{inconsolata}
\usepackage{amsthm}
\usepackage{multirow}

\newcommand{\thickcline}{\noalign{\global\arrayrulewidth=0.4mm}\cline{2-13}\noalign{\global\arrayrulewidth=0.4pt}}

\title{Enhancing In-Context Learning via Implicit Demonstration Augmentation}

\author{
Xiaoling Zhou$^1$, Wei Ye$^1$\footnotemark[1]~,~Yidong Wang$^1$, Chaoya Jiang$^1$, {Zhemg Lee$^{2}$}, \\ {\bf Rui Xie$^1$}, {\bf Shikun Zhang$^1$\footnotemark[1]}\\
$^1$National Engineering Research Center for Software Engineering, Peking University, China\\
$^2$Tianjin University, Tianjin, China \\ 
\texttt{xiaolingzhou@stu.pku.edu.cn,}
\texttt{\{wye,zhangsk\}@pku.edu.cn}
}

\begin{document}
\maketitle
\renewcommand{\thefootnote}{\fnsymbol{footnote}}
\footnotetext[1]{Corresponding authors.}
\renewcommand{\thefootnote}{\arabic{footnote}}

\begin{abstract}
% falls short of supervised fine-tuning, 
% the outcomes
% The emergence of in-context learning (ICL) enables large pre-trained language models to predict labels for unseen inputs without updating parameters. While promising, its performance remains unsatisfactory and  
% frequently exhibits high variance. 
% To address these challenges, this study introduces an \textbf{I}mplicit \textbf{D}emonstration \textbf{A}ugmentation-based \textbf{ICL} (\textbf{IDAICL}) method, which enriches the demonstrations by leveraging semantic directions extracted from the deep feature distribution of the demonstration examples. Importantly, considering the number of augmented copies approaching infinity,
% % To augment more data, we 
% we derive a novel prediction function, termed IDA-Softmax, which incorporates two modulating factors 
% % corresponding to the statistical properties of the input distribution
% to calibrate the original predictions. As a consequence, instead of explicitly augmenting the demonstrations, we can directly utilize IDA-Softmax for making predictions, leading to a highly efficient approach. Extensive experiments demonstrate that our method significantly improves the average and worst-case accuracy across diverse PLMs and tasks. Moreover, it notably reduces performance variance among varying demonstrations, permutations, and templates, as well as displaying capability in addressing imbalanced class distributions.
% as a new prediction function, it seamlessly integrates with other ICL technologies to further enhance performance.

The emergence of in-context learning (ICL) enables large pre-trained language models (PLMs) to make predictions for unseen inputs without updating parameters. Despite its potential, ICL's effectiveness heavily relies on the quality, quantity, and permutation of demonstrations, commonly leading to suboptimal and unstable performance. In this paper, we tackle this challenge for the first time from the perspective of demonstration augmentation. Specifically, we start with enriching representations of demonstrations by leveraging their deep feature distribution. We then theoretically reveal that when the number of augmented copies approaches infinity, the augmentation is approximately equal to a novel logit calibration mechanism integrated with specific statistical properties. This insight results in a simple yet highly efficient method that significantly improves the average and worst-case accuracy across diverse PLMs and tasks. Moreover, our method effectively reduces performance variance among varying demonstrations, permutations, and templates, and displays the capability to address imbalanced class distributions.

\end{abstract}

\section{Introduction}

Large pre-trained language models (PLMs) have showcased exceptional abilities in in-context learning (ICL)~\cite{brown2020language,wang2023investigating,rubin-etal-2022-learning}, which assists the model in discerning the underlying patterns within demonstrations and make more accurate predictions~\cite{chan2022data,wu-etal-2023-self}. 
As a new paradigm, ICL offers compelling advantages, allowing for natural language interaction with PLMs~\cite{wei2022chain,yang2023supervised}, as well as reduced computational costs~\cite{li2023context,rubin-etal-2022-learning}. 

While promising, ICL's performance is highly dependent on provided demonstrations and templates~\cite{liu-etal-2022-makes,zhang-etal-2022-active,sorensen-etal-2022-information}, resulting in subpar and unstable performance. This promotes research aimed at improving the quality~\cite{rubin-etal-2022-learning,li-etal-2023-unified}, quantity~\cite{li2023context,choi2022prompt}, and permutations~\cite{lu-etal-2022-fantastically,tang2023found} of demonstrations. Other research avenues include prediction adjustment~\cite{zhao2021calibrate,han2022prototypical,fei-etal-2023-mitigating} and learning process design (e.g., channel models~\cite{min-etal-2022-noisy} and meta-training frameworks~\cite{min-etal-2022-metaicl}). Despite ongoing efforts, ICL still struggles with efficiently and reliably capturing sufficient knowledge from context, leaving performance stability as a persistent bottleneck.

% Despite promising results, large PLMs still face challenges in efficiently and reliably acquiring adequate knowledge from the context, resulting in large performance variability. 

% Despite promising results, 

% Despite their promise, there remain unanswered questions that warrant further investigation. Primarily, text models confront inherent limitations concerning input text length, and their efficiency diminishes as the number of demonstration examples increases. Consequently, this poses a challenge for PLMs to efficiently and reliably acquire sufficient knowledge from the context~\cite{gao2021limitations,hao2022structured}, resulting in considerable performance variance and suboptimal worst-case accuracy~\cite{zhao2021calibrate,han2022prototypical}. Moreover, the scarcity of annotated data~\cite{wang-etal-2023-self-instruct} underscores the necessity to enhance performance with limited demonstration examples.
% to incorporate substantial knowledge through demonstrations to PLMs~\cite{gao2021limitations,hao2022structured}. 
% without constraints. 
% lags behind supervised fine-tuning, as well as 

\begin{figure}[t] 
\centering
% \vspace{-0.02in}
\includegraphics[width=0.48\textwidth]{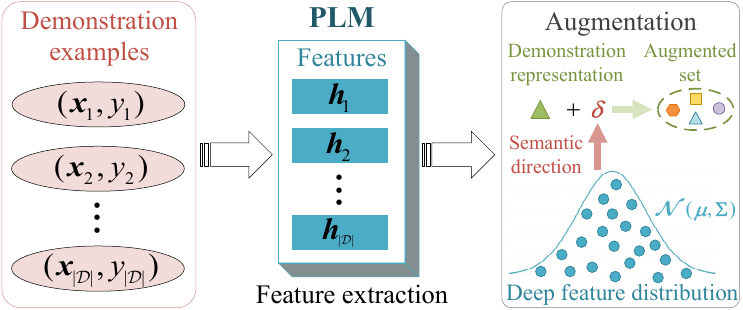}
% \includegraphics[width=0.48\textwidth]{pic/fig11213111.pdf}
% \vspace{-0.02in}
\caption{Illustration for demonstration augmentation using semantic directions (vectors) sampled from the deep feature distribution of demonstration examples.}
\label{demonaug}
% \vspace{-0.03in}
% \vspace{-0.16in}
\end{figure}

In this study, we propose enriching contextual knowledge for PLMs by augmenting demonstrations. We first attempt to enhance the representation of demonstrations by transforming them along semantic directions sampled from the deep feature space of demonstration examples, as depicted in Figure~\ref{demonaug}. This operation stems from the observation that the deep features in a network are usually
linearized~\cite{bengio2013better,cheung2020modals,cho2016noisy}, implying the existence of numerous semantic directions within the deep feature space, hence potentially enabling us to incorporate richer contextual knowledge without extending input length. From this novel perspective, we theoretically prove that when the number of augmented pieces approaches infinity, its effect approximately equals a logit adjustment operation. Specifically, we derive a refined Softmax function that integrates the statistical properties of demonstrations.
Consequently, rather than explicitly executing the augmentation procedure, we can efficiently conduct implicit demonstration augmentation using the derived prediction function, obtaining an improved ICL method with theoretical guidance. 

% only a minimal increase in computational overhead when compared to Vanilla ICL. 

% and large PLMs 
% . We attribute
% the strong performance of IDAICL to their
% stability: they have lower variance and signifi-
% cantly higher worst-case accuracy then their direct
% counterparts over different verbalizers and seeds.
% improves prediction accuracy and 

% \footnote{Code is available at \url{https://github.com/xiaolingzhou98/IDAICL}.}

We conduct extensive experiments across seven PLMs and various classification tasks. The empirical results demonstrate that our approach remarkably enhances prediction accuracy and reduces performance variability across different demonstrations, permutations, and templates. 
Notably, our method is straightforward, effective, and generalizable, enabling seamless integration with other ICL methods to enhance their performance.

Our contributions can be summarized as follows:
\begin{itemize}[itemsep=2pt, topsep=2pt,parsep=2pt]
    \item We introduce \textbf{I}mplicit \textbf{D}emonstration \textbf{A}ugmentation-based \textbf{ICL} (IDAICL), a pioneering work that incorporates demonstration augmentation into ICL. Instead of solely enhancing demonstration quality, quantity, or order, our method explores context augmentation within the deep feature space, offering a new perspective to enrich demonstrations bypassing input length limitations.
    % conducting context augmentation in deep feature space provides another novel perspective to enrich demonstration 

    \item We theoretically establish that as the number of augmented pieces approaches infinity, our augmentation strategy approximates a logit-adjusted prediction function that integrates statistical properties derived from 
    the input data distribution. Equipped with this function, IDAICL provides a straightforward yet theory-guided solution to enhance ICL.
    
      \item Extensive experiments conducted across diverse tasks and PLMs conclusively illustrate that IDAICL considerably improves average and worst-case accuracy compared to existing ICL methods. Moreover, it effectively enhances performance stability.

% the seamless integration of IDAICL with other ICL methods, 
\end{itemize}

\section{Background and Related Work}
\subsection{In-Context Learning}
Brown et al.~\shortcite{brown2020language} showcased the ICL capability of PLMs, wherein PLMs generate predictions solely based on a concatenation of training examples for few-shot learning without updating parameters. 
% Brown et al.~\shortcite{brown2020language} proposed 
Subsequent studies~\cite{holtzman-etal-2021-surface,min-etal-2022-noisy,min-etal-2022-metaicl} have developed this approach, yielding promising outcomes across various tasks. 
Nevertheless, recent research has uncovered certain limitations. To begin with, the volume of input knowledge for each query is constrained by the maximum input length of PLMs~\cite{hao2022structured}, and the computational cost increases as the number of demonstrations grows~\cite{li2023context}, making it challenging to integrate significant knowledge from demonstrations to PLMs. Additionally, ICL's performance is sensitive to the input of PLMs~\cite{davison-etal-2019-commonsense, jiang2020can}, 
thus exhibiting high variance and poor worst-case accuracy~\cite{perez2021true,lu-etal-2022-fantastically}. 

Researchers have explored various techniques to address the biases and instability of ICL. These techniques encompass learning process design~\cite{min-etal-2022-noisy,min-etal-2022-metaicl}, demonstration retrieval~\cite{rubin-etal-2022-learning,zhang-etal-2022-active}, prompt engineering~\cite{sorensen-etal-2022-information,lu-etal-2022-fantastically}, and prediction calibration~\cite{zhao2021calibrate,fei-etal-2023-mitigating}. However, these methods have yet to fully address the issue of severely limited knowledge transfer from demonstrations to large PLMs. 
% knowledge transfer from demonstrations to large PLMs.
% meta-training frameworks~\cite{min-etal-2022-metaicl}, channel models~\cite{min-etal-2022-noisy}, 
% This study builds upon the fundamental concept of ICL by conditioning the PLMs on training examples~\cite{brown2020language}.

% improved prediction techniques~\cite{zhao2021calibrate, holtzman-etal-2021-surface, han2022prototypical}. For example, Zhao et al.~\shortcite{zhao2021calibrate} calibrated the probabilities of a PLM using a learned affine transformation, while Holtzman et al.~\shortcite{holtzman-etal-2021-surface} introduced domain conditional pointwise
% mutual information, which is an alternative scoring function that directly compensates for surface form competition. 
%  

\subsection{Data Augmentation}

\begin{figure*}[t] 
\centering
% \vspace{-0.1in}
\includegraphics[width=1\textwidth]{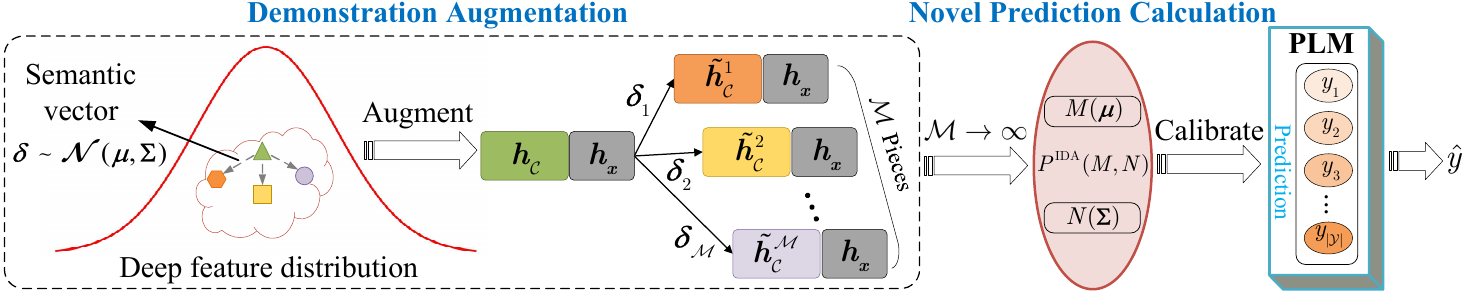}
% \vspace{-0.12in}
\caption{An overview of IDAICL: For each contextual input, our goal is to augment the deep feature of demonstrations for $\mathcal{M}$ pieces, using semantic vectors $\boldsymbol{\delta}$ drawn from the deep feature distribution $\mathcal{N}(\boldsymbol{\mu},\boldsymbol{\Sigma})$ of demonstration examples linked to all queries. When $\mathcal{M}$ approaches infinity, we derive a novel prediction function, which incorporates two modulating factors: $M(\boldsymbol{\mu})$ and $N(\boldsymbol{\Sigma})$, to calibrate the original predictions. 
}
\label{kuangjia}
% \vspace{-0.02in}
% \vspace{-0.16in}
\end{figure*}

Data augmentation~\cite{chen-etal-2023-empirical}, which involves artificially creating training data through transformations, is a well-established research area in machine learning. Although data augmentation techniques have undergone extensive exploration in diverse machine learning domains~\cite{maharana2022review,shorten2019survey}, applying them to text data poses challenges due to the complexity of preserving labels during textual transformations~\cite{kobayashi-2018-contextual}.
Nonetheless, data augmentations in the latent space, such as adversarial training~\cite{zhang2022adversarial,zhu2019freelb,cheng-etal-2020-advaug}, interpolation~\cite{chen-etal-2022-doublemix, wu-etal-2022-text}, and generative techniques~\cite{li-etal-2022-generative,malandrakis-etal-2019-controlled}, have demonstrated notable enhancements when applied alongside large PLMs. 

Recently, Wang et al.~\shortcite{r8} introduced the concept of implicit data augmentation in the context of image classification.
This approach involves transforming training data within the deep feature space and boils down to the optimization of a novel robust loss function.
% within image classification scenarios. 
% Recently, Wang et al.~\shortcite{r8} introduced the concept of implicit data augmentation in the realm of image classification, which augments data in the deep feature space and boils down to the optimization of a novel robust loss funtion. 
% which augments training data in the deep feature space. Additionally, instead of conducting the augmentation procedure explicitly, this process is only achieved through 
% data augmentation is exclusively achieved through 
% the optimization of a robust loss function. 
Subsequent studies~\cite{r79,r9,zhou2023implicit} for image classification tasks have further improved upon this approach. This study introduces an algorithm for implicitly augmenting demonstrations within the realm of ICL.
% This study introduces an algorithm for 
% the implicit augmentation of demonstrations in the context of ICL. 
% which serves as a novel prediction objective.
% This study focuses on augmenting demonstrations with semantic directions within the continuous feature space.

\section{Methodology}

\subsection{In-Context Learning with PLMs}

Considering a PLM $\mathcal{G}$, 
this study focuses on the following task: given a query input text $\boldsymbol{x}$ and a candidate answer set $\mathcal{Y}\!=\!\{y_1, y_2, \cdots, y_{|\mathcal{Y}|}\}$,
we aim to predict the answer $\hat{y}$ based on $m$ demonstration examples $\mathcal{C}\!=\!\{{c}_{1}, c_{2}, \cdots, c_{m}\}$, where each $c_{i}$ represents a training example $(\boldsymbol{x}_{i},y_{i})$ after template formulation and $m$ denotes the quantity of demonstration examples for each test sample. Formally, give a model $\mathcal{G}$, we first compute the probability of each answer $y_j$:
\begin{equation}
P_{\mathcal{G}}\left(y_j \mid \mathcal{C}, \boldsymbol{x}\right).
\end{equation}
Subsequently, the ultimate prediction $\hat{y}$, characterized by the highest probability is chosen from the candidate answer set $\mathcal{Y}$:

\begin{equation}
\hat{y}=\arg\max_{y_{j}\in\mathcal{Y}}P_{\mathcal{G}}\left(y_j \mid \mathcal{C}, \boldsymbol{x}\right).
\end{equation}

To simplify, the contextual input is denoted as $\boldsymbol{\tilde{x}}\!=\![\mathcal{C},\boldsymbol{x}]$ in the subsequent text. Then, the probability of answer $y_j$, represented as $P_{\mathcal{G}}(y_j|\boldsymbol{\tilde{x}})$, is computed using the Softmax function\footnote{We begin by examining situations in which the answer comprises a single token, and our subsequent analysis is equally applicable to scenarios involving multiple tokens.}:
\begin{equation}
    P_{\mathcal{G}}(y_j|\boldsymbol{\tilde{x}}) :=P_{\mathcal{G}}(y_j|\boldsymbol{h}_{\boldsymbol{\tilde{x}}})= \frac{e^{\boldsymbol{w}_{y_j}^{T}\boldsymbol{h}_{\tilde{\boldsymbol{x}}}+b_{y_j}}}{\sum_{k}e^{\boldsymbol{w}_{k}^{T}\boldsymbol{h}_{\tilde{\boldsymbol{x}}}+b_{k}}},
\end{equation}
where $\boldsymbol{h}_{\tilde{\boldsymbol{x}}}=\mathcal{G}(\tilde{\boldsymbol{x}})$ signifies the hidden state of the last block at the final position for $\tilde{\boldsymbol{x}}$. $\boldsymbol{w}_{k}$ and $b_{k}$ are the weight vector and bias corresponding to the final fully connected layer for the $k$-th token.

\subsection{Demonstration Augmentation}
% Previous research~\cite{liu-etal-2022-makes} has revealed that increasing the volume of knowledge conveyed to PLMs leads to improved performance. 
% the representations  
Recognizing the established efficacy of data augmentation in machine learning~\cite{feng-etal-2021-survey}, this study investigates demonstration augmentation and 
suggests enhancing the deep features of demonstrations by transforming them along semantic directions sampled from the deep feature space of demonstration examples. This strategy is motivated by the intriguing observation
that the deep features in networks are often linearized~\cite{bengio2013better,chen2022cross}. Building on this observation, we hypothesize that $\boldsymbol{h}_{\tilde{\boldsymbol{x}}}$ lies within the subspace spanned by $\boldsymbol{h}_{\mathcal{C}}$ and $\boldsymbol{h}_{{\boldsymbol{x}}}$: $\boldsymbol{h}_{\tilde{\boldsymbol{x}}}=\alpha\boldsymbol{h}_{\mathcal{C}}+\beta\boldsymbol{h}_{\boldsymbol{x}}$, where $\boldsymbol{h}_{\mathcal{C}}$ and $\boldsymbol{h}_{{\boldsymbol{x}}}$ represent 
the components of $\boldsymbol{h}_{\tilde{\boldsymbol{x}}}$ 
linked respectively to the demonstrations and the query. The necessity of this assumption stems from intricate relationships among token representations and the exclusive augmentation of the component related to demonstrations. Notably, this decomposition is not necessary in practical applications. In the subsequent text, we directly refer to $\alpha\boldsymbol{h}_{\mathcal{C}}$ and $\beta\boldsymbol{h}_{\boldsymbol{x}}$ as $\boldsymbol{h}_{\mathcal{C}}$ and $\boldsymbol{h}_{\boldsymbol{x}}$.
% are linearly independent, where $\boldsymbol{h}_{\mathcal{C}}$ and $\boldsymbol{h}_{{\boldsymbol{x}}}$ denote 
% we expressed c as a.Given the intricate relationships among token representations and the fact that $\boldsymbol{h}_{\tilde{\boldsymbol{x}}}$ encompasses information from both demonstrations and the test sample, we ideally divide $\boldsymbol{h}_{\tilde{\boldsymbol{x}}}$ into two components:
% % consider it a combination of 
% information of demonstrations, $\boldsymbol{h}_{\mathcal{C}}$, and that of the test sample, $\boldsymbol{h}_{\boldsymbol{x}}$, expressed as $\boldsymbol{h}_{\tilde{\boldsymbol{x}}}\!:=\!\boldsymbol{h}_{\mathcal{C}}\!+\!\boldsymbol{h}_{\boldsymbol{x}}$.
% This expression is solely for the convenience of deriving
% the loss function after augmenting demonstrations and is not
% required in practical applications.
% This expression is sorely for the convenience of deriving the loss  

To augment $\boldsymbol{h}_{\mathcal{C}}$, we randomly sample vectors from the deep feature space of demonstrations. In particular, vectors are drawn from a multivariate normal distribution 
$\mathcal{N}(\boldsymbol{\mu},\boldsymbol{\Sigma})$, where $\boldsymbol{\mu}$ and $\boldsymbol{\Sigma}$ denote the feature mean and covariance matrix. These statistical properties are estimated from the deep features of the demonstration set $\mathcal{D}$, which includes demonstration examples linked to all queries. 
% estimated from the deep features of  $\mathcal{D}$ which contains demonstrations linked to all queries. 
The feature mean $\boldsymbol{\mu}$ is computed as
\begin{equation}
    \boldsymbol{\mu} = \frac{1}{|\mathcal{D}|}\sum\nolimits_{i=1}^{|\mathcal{D}|}\boldsymbol{h}_{i},
\end{equation}
where $\boldsymbol{h}_{i} = \mathcal{G}(c_{i})$ represents the hidden state of the last block at the final position for the $i$-th demonstration example $c_{i}$ in $\mathcal{D}$, and $|\mathcal{D}|$ denotes the size of $\mathcal{D}$.
% which generally equals to $|\mathcal{D}^{\text{te}}|*m$. $\mathcal{D}^{\text{te}}$ refers to the test set. 
The covariance matrix $\boldsymbol{\Sigma}$ is computed as
\begin{equation}
    \boldsymbol{\Sigma} = \frac{1}{|\mathcal{D}|}{\sum\nolimits_{i=1}^{|\mathcal{D}|}(\boldsymbol{h}_{i}-\boldsymbol{\mu})^{T}(\boldsymbol{h}_{i}-\boldsymbol{\mu})}.
\end{equation}
% In our implementation, we present two approaches for calculating mean and covariance values. The first method entails precomputing the mean and covariance before performing inference. The second approach utilizes an online estimation method during the inference process. For a more comprehensive discussion, please consult Appendix~\ref{statis}.
% the mean and covariance matrix values are dynamically computed online by aggregating statistics from all mini-batches, as elaborated in Section~\ref{statis}.

Subsequently, ${\boldsymbol{h}}_{\mathcal{C}}$ is shifted in the extracted semantic vectors, resulting in augmented features, $\tilde{\boldsymbol{h}}_{\mathcal{C}}$, which follows
\begin{equation}
% \boldsymbol{h}_{\mathcal{C}}+\boldsymbol{\delta}  = 
\tilde{\boldsymbol{h}}_{\mathcal{C}}  \sim \mathcal{N}\left({\boldsymbol{h}}_{\mathcal{C}}+\lambda\boldsymbol{\mu}, \lambda\boldsymbol{\Sigma}\right),
\end{equation}
where $\lambda$ refers to a positive coefficient controlling the strength of semantic augmentation. In real-world applications, it can be directly assigned a value of 0.5. Sensitivity tests for $\lambda$ are discussed in Section~\ref{secsense11}.
% When dealing with restricted training data, it becomes practical to leverage the statistics derived from the complete training set.
% % , which entails enhancing demonstrations of the entire training space. 
% Alternatively, estimation can be conducted with a subset of training samples, that is, the demonstration set. 
% Besides pre-inference calculation of these statistical properties, we also introduce an online estimation algorithm for computing $\boldsymbol{\mu}$ and $\boldsymbol{\Sigma}$ in Section~\ref{statis} of the Appendix.
% has the potential to yield superior results.

\subsection{Novel Prediction Function}
Selecting the answer with the highest probability is equivalent to favoring the answer with the lowest inverse probability. Therefore, the prediction can be determined by
\begin{equation}
\hat{y}=\arg\min_{y_{j}\in\mathcal{Y}}P_{\mathcal{G}}\left(y_j \mid  \boldsymbol{h}_{\tilde{\boldsymbol{x}}}\right)^{-1}.
\end{equation}
Assume that each $\boldsymbol{h}_{\mathcal{C}}$ is augmented for $\mathcal{M}$ times, resulting in an
augmented demonstration feature set $\{\tilde{\boldsymbol{h}}_{\mathcal{C}}^{1}, \cdots, \tilde{\boldsymbol{h}}_{\mathcal{C}}^{\mathcal{M}}\}$ with size $\mathcal{M}$. Here, $\tilde{\boldsymbol{h}}_{\mathcal{C}}^{i}$ represents the $i$-th augmented
feature for $\boldsymbol{h}_{\mathcal{C}}$. Then, the final prediction for the query $\boldsymbol{x}$ depends on all augmented features of $\boldsymbol{h}_{\mathcal{C}}$ and can be expressed as
% \begin{equation}
%     \overline{\ell}_{y_j}(\boldsymbol{\tilde{x}}) = 
%     % \overline{P}_{\mathcal{G}}(y_{j}|\mathcal{C},\boldsymbol{x}) = 
%     \frac{1}{\mathcal{M}}{\sum_{i=1}^{\mathcal{M}}\sum_{v\in y_{j}}-\log[P_{\mathcal{G}}(v|\tilde{\boldsymbol{h}}_{\mathcal{C}}^{i},\boldsymbol{h}_{\boldsymbol{x}})]},
% \end{equation}
% \begin{equation}
%     \overline{\ell}_{y_j}(\boldsymbol{\tilde{x}}) = 
%     % \overline{P}_{\mathcal{G}}(y_{j}|\mathcal{C},\boldsymbol{x}) = 
%     \frac{1}{\mathcal{M}}{\sum\nolimits_{i=1}^{\mathcal{M}}-\log[P_{\mathcal{G}}(y_j|\tilde{\boldsymbol{h}}_{\mathcal{C}}^{i},\boldsymbol{h}_{\boldsymbol{x}})]},
% \end{equation}
% \begin{equation}
%     \overline{P}_{y_j}(\boldsymbol{\tilde{x}}) = 
%     % \overline{P}_{\mathcal{G}}(y_{j}|\mathcal{C},\boldsymbol{x}) = 
%     \frac{1}{\mathcal{M}}{\sum\nolimits_{i=1}^{\mathcal{M}}P_{\mathcal{G}}(y_j|\tilde{\boldsymbol{h}}_{\mathcal{C}}^{i},\boldsymbol{h}_{\boldsymbol{x}})},
% \end{equation}
\begin{equation}
    {P}_{y_j}^{\mathcal{M}}(\boldsymbol{\tilde{x}}) = 
    % \overline{P}_{\mathcal{G}}(y_{j}|\mathcal{C},\boldsymbol{x}) = 
    \frac{1}{\mathcal{M}}{\sum\nolimits_{i=1}^{\mathcal{M}} P_{\mathcal{G}}(y_j|\tilde{\boldsymbol{h}}_{\mathcal{C}}^{i},\boldsymbol{h}_{\boldsymbol{x}})^{-1}},
\end{equation}
% \begin{equation}
% \hat{y}(\tilde{\boldsymbol{x}})={\arg \min_{y_{j}\in\mathcal{Y}} \overline{\ell}_{y_j}(\boldsymbol{\tilde{x}})}.
% \end{equation}
% \begin{equation}
% \hat{y}(\tilde{\boldsymbol{x}})={\arg \max_{y_{j}\in\mathcal{Y}} \overline{P}_{y_j}(\boldsymbol{\tilde{x}})}.
% \end{equation}
\begin{equation}
\hat{y}={\arg \min_{y_{j}\in\mathcal{Y}} {P}_{y_j}^{\mathcal{M}}(\boldsymbol{\tilde{x}})}.
\end{equation}

% However, expanding the feature set $\mathcal{M}$ times would lead to increased computational demands, especially when 
% $\mathcal{M}$ is large. To improve efficiency while generating more data, we 
% explore augmenting infinite times and 
Given that the performance of ICL benefits from an increased number of demonstration instances~\cite{liu-etal-2022-makes,wu-etal-2023-self}, 
we explore the scenario of augmenting an infinite number of times for the deep representation of demonstrations.
% of each test sample. 
Subsequently, an easily computable surrogate for the expected prediction can be derived, resulting in a highly efficient implementation. 
% and then derived an easy-to-compute surrogacy for the expected prediction, which thus leads to a highly efficient implementation. 
The whole pipeline of IDAICL is depicted in Figure~\ref{kuangjia}.

As $\mathcal{M} \rightarrow \infty$, on the basis of the aforementioned decomposition of $\boldsymbol{h}_{\tilde{\boldsymbol{x}}}$, the expected prediction for answer $y_{j}$ (denoted as $P^{\infty}_{y_j}$) within the augmented feature set can be expressed as follows:
% \begin{equation}
% \resizebox{1\linewidth}{!}{$
% \begin{aligned}
%     \ell_{y_j}^{\infty}(\boldsymbol{\tilde{x}}) = \mathbb{E}_{\boldsymbol{\tilde{h}}_{\mathcal{C}}}[\sum_{v\in y_j}-\log(\frac{e^{\boldsymbol{w}_{v}^T [\boldsymbol{\tilde{h}}_{\mathcal{C}};\boldsymbol{h}_{\boldsymbol{x}}]+b_{v}}}{\sum_{k} e^{\boldsymbol{w}_k^T [\boldsymbol{\tilde{h}}_{\mathcal{C}};\boldsymbol{h}_{\boldsymbol{x}}]+b_k}})].
% \end{aligned}$}
% \label{ave_loss}
% \end{equation}
% \begin{equation}
% % \resizebox{1\linewidth}{!}{$
% \begin{aligned}
%     \ell_{y_j}^{\infty}(\boldsymbol{\tilde{x}}) := \mathbb{E}_{\boldsymbol{\tilde{h}}_{\mathcal{C}}}[-\log(\frac{e^{\boldsymbol{w}_{y_j}^T (\boldsymbol{\tilde{h}}_{\mathcal{C}}+\boldsymbol{h}_{\boldsymbol{x}})+b_{y_j}}}{\sum_{k} e^{\boldsymbol{w}_k^T (\boldsymbol{\tilde{h}}_{\mathcal{C}}+\boldsymbol{h}_{\boldsymbol{x}})+b_k}})].
% \end{aligned}
% % $}
% \label{ave_loss}
% \end{equation}
\begin{equation}
\begin{aligned}
    P_{y_j}^{\infty}(\boldsymbol{\tilde{x}})\!=\!\mathbb{E}_{\boldsymbol{\tilde{h}}_{\mathcal{C}}}[{\sum_{k} e^{\Delta\boldsymbol{w}_{k,y_j}^T (\boldsymbol{\tilde{h}}_{\mathcal{C}}+\boldsymbol{h}_{\boldsymbol{x}})+\Delta b_{k,y_j}}}],
\end{aligned}
\label{ave_loss}
\end{equation}
where $\Delta \boldsymbol{w}_{k,y_j} = \boldsymbol{w}_k-\boldsymbol{w}_{y_j}$ and $\Delta b_{k,y_j} = b_k-b_{y_j}$.

However, accurately calculating $P^{\infty}_{y_j}$
% the aforementioned expected loss 
is challenging. Alternatively, we proceed to derive a surrogate calculation for it. Applying the linearity of expectation, Eq.~(\ref{ave_loss}) can be expressed as:
\begin{equation}
\begin{aligned}
    P_{y_j}^{\infty}(\boldsymbol{\tilde{x}})\!=\!\sum_{k} \mathrm{E}_{\tilde{\boldsymbol{h}}_{\mathcal{C}}}[e^{\Delta \boldsymbol{w}_{k,y_j}^{T} (\tilde{\boldsymbol{h}}_{\mathcal{C}}+\boldsymbol{h}_{\boldsymbol{x}})+\Delta b_{k,y_j}}].
\end{aligned}
\label{convex}
\end{equation}
Given that $\boldsymbol{\tilde{h}}_{\mathcal{C}}$ is a Gaussian random variable conforming to $\mathcal{N}\left({\boldsymbol{h}}_{\mathcal{C}}+\lambda\boldsymbol{\mu}, \lambda\boldsymbol{\Sigma}\right)$, we know that $\Delta \boldsymbol{w}_{k,y_j}^{T} \tilde{\boldsymbol{h}}_{\mathcal{C}}$ follows the multivariate normal distribution: $\mathcal{N}(\Delta \boldsymbol{w}_{k,y_j}^{T} \left({\boldsymbol{h}}_{\mathcal{C}}+\lambda\boldsymbol{\mu}\right), \lambda\Delta \boldsymbol{w}_{k,y_j}^{T} \boldsymbol{\Sigma} \Delta \boldsymbol{w}_{k,y_j})$. Then, utilizing the moment-generating function
\begin{equation}
    \mathbb{E}[e^{{t}{X}}] = e^{{t}{\mu}+\frac{1}{2}{t}^{2}\sigma^{2}}, X \sim \mathcal{N}(\mu,\sigma^{2}),
\end{equation}
Eq.~(\ref{convex}) can be derived as
\begin{equation}
\begin{aligned}
    P_{y_j}^{\infty}(\boldsymbol{\tilde{x}})\!=\!\sum_{k} M_{k,y_j}N_{k,y_j}e^{\Delta\boldsymbol{w}_{k,y_j}^{T}(\boldsymbol{h}_{\mathcal{C}}+\boldsymbol{h}_{\boldsymbol{x}})+\Delta b_{k,y_j}}, 
\end{aligned}
\label{zuizhong}
\end{equation}
% \begin{equation}
% \begin{aligned}
%     P_{y_j}^{\infty}(\boldsymbol{\tilde{x}})\!=\!\sum_{k} e^{\Delta\boldsymbol{w}_{k,y_j}^{T}\boldsymbol{h}_{\boldsymbol{x}}+\Delta b_{k,y_j}+M_{k,y_j}+N_{k,y_j}}, 
% \end{aligned}
% \label{zuizhong}
% \end{equation}
where $M_{k,y_j} = \exp(\lambda{\Delta\boldsymbol{w}_{k,y_j}^{T}\boldsymbol{\mu}})$ and $N_{k,y_j} = \exp({{\frac{\lambda}{2}}\Delta\boldsymbol{w}_{k,y_j}^{T} \boldsymbol{\Sigma}\Delta\boldsymbol{w}_{k,y_j}})$. 

\begin{table*}[t]
% \small
% \vspace{-0.05in}
\centering
\resizebox{1\linewidth}{!}{
\begin{tabular}{l|l|c|cccccccccc}
\toprule[1.5pt]
\rowcolor{gray!30}
        {PLM} & Method  & m     & SST-2 & SST-5 & MR & CR & Amazon & Subj & TREC & DBPedia & AGNews & CB \\ \midrule\midrule
% \midrule \midrule
     \multirow{12}*{\rotatebox{90}{GPT-2 0.8B}} 
     & Vanilla ICL & \multirow{2}{*}{4}             &  ${57.6}_{\textcolor{blue}{7.1}}$     &  ${30.4}_{\textcolor{blue}{6.3}}$     & ${59.3}_{\textcolor{blue}{6.5}}$   & ${56.8}_{\textcolor{blue}{8.4}}$   & $32.7_{\textcolor{blue}{8.5}}$     & ${57.6}_{\textcolor{blue}{5.4}}$     & ${34.9}_{\textcolor{blue}{10.3}}$     & ${40.5}_{\textcolor{blue}{7.2}}$        & ${44.5}_{\textcolor{blue}{7.9}}$       & ${35.1}_{\textcolor{blue}{9.3}}$   \\ 
& IDAICL &         &  $86.4_{\textcolor{blue}{1.4}}$     & $38.3_{\textcolor{blue}{2.9}}$      & $82.2_{\textcolor{blue}{2.3}}$   & $78.4_{\textcolor{blue}{0.7}}$   & $46.7_{\textcolor{blue}{3.5}}$     & $77.0_{\textcolor{blue}{2.3}}$     & $47.5_{\textcolor{blue}{2.0}}$     & $81.3_{\textcolor{blue}{1.8}}$        & $73.9_{\textcolor{blue}{2.4}}$       & $41.5_{\textcolor{blue}{2.0}}$   \\ \cline{2-13}

& Vanilla ICL & \multirow{2}{*}{8}             &  ${69.7}_{\textcolor{blue}{9.0}}$     &  ${32.4}_{\textcolor{blue}{8.6}}$     & ${63.9}_{\textcolor{blue}{7.7}}$   & ${60.8}_{\textcolor{blue}{8.1}}$   & $34.1_{\textcolor{blue}{6.2}}$     & ${59.7}_{\textcolor{blue}{8.7}}$     & ${40.4}_{\textcolor{blue}{6.3}}$     & ${62.6}_{\textcolor{blue}{13.6}}$        & ${49.2}_{\textcolor{blue}{8.4}}$       & ${38.8}_{\textcolor{blue}{7.6}}$  \\ 
& IDAICL &         &  $88.0_{\textcolor{blue}{2.3}}$     & $39.6_{\textcolor{blue}{1.9}}$      & $84.9_{\textcolor{blue}{2.4}}$   & ${85.6}_{\textcolor{blue}{2.5}}$   &   $47.9_{\textcolor{blue}{2.6}}$   & $79.9_{\textcolor{blue}{0.8}}$      &    ${50.3}_{\textcolor{blue}{3.3}}$  &  ${86.5}_{\textcolor{blue}{2.9}}$       &     $76.8_{\textcolor{blue}{1.7}}$   & $43.3_{\textcolor{blue}{3.4}}$   \\ \cline{2-13}
     &  Vanilla ICL  & \multirow{2}{*}{12}           &  ${74.7}_{\textcolor{blue}{8.3}}$     & ${33.7}_{\textcolor{blue}{7.6}}$      &   ${64.4}_{\textcolor{blue}{9.4}}$  &  ${68.7}_{\textcolor{blue}{9.7}}$  &  ${36.0}_{\textcolor{blue}{6.6}}$    &  ${60.7}_{\textcolor{blue}{7.7}}$    &  ${40.5}_{\textcolor{blue}{7.8}}$    &     ${64.5}_{\textcolor{blue}{5.4}}$    &    ${51.1}_{\textcolor{blue}{8.0}}$    & ${40.4}_{\textcolor{blue}{8.5}}$   \\ 
& IDAICL    &      &   $88.5_{\textcolor{blue}{2.1}}$    & $40.1_{\textcolor{blue}{2.7}}$      & $85.2_{\textcolor{blue}{3.1}}$   &  $86.8_{\textcolor{blue}{1.4}}$  &  $49.6_{\textcolor{blue}{2.2}}$    & $80.4_{\textcolor{blue}{2.1}}$     &  $51.4_{\textcolor{blue}{1.6}}$    &   $87.3_{\textcolor{blue}{2.7}}$      &     ${77.9}_{\textcolor{blue}{2.0}}$   &  $44.6_{\textcolor{blue}{2.2}}$  \\ \thickcline
& MetaICL    & \multirow{2}{*}{12}     &   ${80.8}_{\textcolor{blue}{6.2}}$    &    ${35.8}_{\textcolor{blue}{4.7}}$   &  ${75.3}_{\textcolor{blue}{5.6}}$  & $77.6_{\textcolor{blue}{8.1}}$   &   $48.9_{\textcolor{blue}{6.7}}$   &  ${73.5}_{\textcolor{blue}{8.8}}$    &   $48.6_{\textcolor{blue}{6.1}}$   &  $80.4_{\textcolor{blue}{7.8}}$       &    ${66.8}_{\textcolor{blue}{0.7}}$    &  ${{43.1}}_{\textcolor{blue}{4.1}}$  \\
& +IDAICL  &    &   ${89.3}_{\textcolor{blue}{1.7}}$    &  ${\underline{42.6}}_{\textcolor{blue}{2.4}}$     & ${85.8}_{\textcolor{blue}{1.7}}$   &  ${87.9}_{\textcolor{blue}{1.5}}$  &  ${\underline{51.7}}_{\textcolor{blue}{0.7}}$    &  ${\underline{82.6}}_{\textcolor{blue}{2.4}}$    & ${53.7}_{\textcolor{blue}{2.5}}$     &   ${\underline{89.4}}_{\textcolor{blue}{4.1}}$      &    $78.3_{\textcolor{blue}{1.1}}$    & ${\boldsymbol{47.9}}_{\textcolor{blue}{2.8}}$   \\ \cline{2-13}
& Channel ICL & \multirow{2}{*}{12} & ${85.2}_{\textcolor{blue}{3.6}}$     &   ${38.4}_{\textcolor{blue}{4.3}}$    & ${80.8}_{\textcolor{blue}{4.7}}$   &  ${82.0}_{\textcolor{blue}{4.6}}$  &  ${43.6}_{\textcolor{blue}{5.1}}$    & ${69.8}_{\textcolor{blue}{9.8}}$     &  ${44.1}_{\textcolor{blue}{8.7}}$    &   ${77.6}_{\textcolor{blue}{12.9}}$      &  ${69.5}_{\textcolor{blue}{6.7}}$    & ${42.4}_{\textcolor{blue}{5.2}}$    \\
& +IDAICL &     &   ${\boldsymbol{90.5}}_{\textcolor{blue}{2.3}}$    & ${41.8}_{\textcolor{blue}{2.7}}$      & ${\boldsymbol{87.7}}_{\textcolor{blue}{1.6}}$   &  ${\boldsymbol{89.5}}_{\textcolor{blue}{1.2}}$  &  ${50.8}_{\textcolor{blue}{2.4}}$    &  ${80.5}_{\textcolor{blue}{0.9}}$    & ${52.9}_{\textcolor{blue}{1.6}}$     & ${87.8}_{\textcolor{blue}{2.4}}$        &  ${\underline{81.0}}_{\textcolor{blue}{2.5}}$      & ${46.3}_{\textcolor{blue}{3.3}}$   \\ 
\cline{2-13}
% SimCSE     &       &       &    &    &      &      &      &         &        &     \\
% IDA+SimCSE &       &       &    &    &      &      &      &         &        &    \\ \hline
% & $k$NN-LM     & ${65.4}_{\textcolor{blue}{7.2}}$       & 
%  $40.6_{\textcolor{blue}{3.5}}$      &   ${56.4}_{\textcolor{blue}{6.8}}$ &  ${69.2}_{\textcolor{blue}{5.9}}$   &  ${57.4}_{\textcolor{blue}{8.1}}$    & $62.4_{\textcolor{blue}{6.1}}$      & $55.9_{\textcolor{blue}{3.1}}$    &   $56.9_{\textcolor{blue}{6.7}}$       &   ${67.0}_{\textcolor{blue}{1.4}}$      & $65.6_{\textcolor{blue}{4.1}}$   \\
% & $k$NN-LM+IDA &   $70.6_{\textcolor{blue}{3.5}}$    & $\underline{47.4}_{\textcolor{blue}{1.8}}$      &  $68.7_{\textcolor{blue}{3.2}}$  & $76.8_{\textcolor{blue}{3.5}}$   &  ${61.9}_{\textcolor{blue}{4.1}}$    &  $69.9_{\textcolor{blue}{3.6}}$    & $58.8_{\textcolor{blue}{2.2}}$     & $66.7_{\textcolor{blue}{3.6}}$        &  $70.8_{\textcolor{blue}{1.2}}$      & $68.5_{\textcolor{blue}{3.5}}$   \\ \cline{2-12}
% % $k$NN-Prompt     & $84.2$       &       &   $78.2$ &  $84.3$   &      &       &     &          &    $78.8$     & $57.2$   \\
% % IDA+$k$NN-Prompt &       &       &    &    &      &      &      &         &        &    \\ \hline
% % GUNDAM     & $76.0$       &   $39.6$    &    &     &      &       &   $55.8$  &          &         &    \\
% % IDA+GUNDAM &       &       &    &    &      &      &      &         &        &    \\ \hline
& EPR       & \multirow{2}{*}{12}       &   ${81.9}_{\textcolor{blue}{2.1}}$    & $39.9_{\textcolor{blue}{1.8}}$      &  $78.1_{\textcolor{blue}{2.4}}$  &  $80.6_{\textcolor{blue}{0.6}}$  &  $49.1_{\textcolor{blue}{2.4}}$    &  $80.1_{\textcolor{blue}{2.2}}$    &   ${\underline{76.2}}_{\textcolor{blue}{1.1}}$   &     ${{87.1}}_{\textcolor{blue}{1.0}}$    &  ${{80.9}}_{\textcolor{blue}{0.8}}$      & $44.8_{\textcolor{blue}{2.3}}$   \\
& +IDAICL  &       &  ${\underline{90.1}}_{\textcolor{blue}{1.1}}$     &  ${\boldsymbol{43.9}}_{\textcolor{blue}{1.2}}$     & ${\underline{86.4}}_{\textcolor{blue}{2.0}}$   & ${\underline{88.6}}_{\textcolor{blue}{0.6}}$   & ${\boldsymbol{52.5}}_{\textcolor{blue}{1.7}}$     & ${\boldsymbol{83.6}}_{\textcolor{blue}{1.0}}$     & ${\boldsymbol{79.1}}_{\textcolor{blue}{0.9}}$     & ${\boldsymbol{90.8}}_{\textcolor{blue}{0.7}}$        &  ${\boldsymbol{83.7}}_{\textcolor{blue}{0.5}}$      & ${\underline{46.7}}_{\textcolor{blue}{2.1}}$   \\ 
 \midrule\midrule
\multirow{12}*{\rotatebox{90}{GPT-2 1.5B}} 
& Vanilla ICL & \multirow{2}{*}{4}             &  ${66.3}_{\textcolor{blue}{8.6}}$     &  ${30.3}_{\textcolor{blue}{8.9}}$     & ${56.5}_{\textcolor{blue}{6.6}}$   & ${53.4}_{\textcolor{blue}{8.1}}$   & $34.7_{\textcolor{blue}{7.5}}$     & ${54.2}_{\textcolor{blue}{5.5}}$     & ${30.8}_{\textcolor{blue}{8.1}}$     & ${61.9}_{\textcolor{blue}{8.7}}$        & ${54.6}_{\textcolor{blue}{9.9}}$       & ${40.8}_{\textcolor{blue}{7.8}}$    \\ 
& IDAICL &          &  $87.4_{\textcolor{blue}{1.5}}$     & $38.8_{\textcolor{blue}{1.7}}$      & $80.9_{\textcolor{blue}{1.2}}$   & $82.1_{\textcolor{blue}{2.1}}$   & $48.1_{\textcolor{blue}{0.6}}$     & $77.8_{\textcolor{blue}{3.0}}$     & $49.5_{\textcolor{blue}{1.9}}$     & $87.4_{\textcolor{blue}{2.6}}$        & $79.2_{\textcolor{blue}{1.8}}$       & $54.1_{\textcolor{blue}{2.7}}$   \\ \cline{2-13}

& Vanilla ICL & \multirow{2}{*}{8}             &  ${57.2}_{\textcolor{blue}{7.0}}$     &  ${30.8}_{\textcolor{blue}{6.1}}$     & ${64.9}_{\textcolor{blue}{8.3}}$   & ${57.6}_{\textcolor{blue}{6.4}}$   & $38.6_{\textcolor{blue}{6.4}}$     & ${57.3}_{\textcolor{blue}{10.3}}$     & ${39.5}_{\textcolor{blue}{5.3}}$     & ${67.4}_{\textcolor{blue}{8.1}}$        & ${56.3}_{\textcolor{blue}{5.4}}$       & ${47.4}_{\textcolor{blue}{5.1}}$  \\ 
& IDAICL &         &  $89.5_{\textcolor{blue}{1.8}}$     & $40.8_{\textcolor{blue}{1.9}}$      & $82.1_{\textcolor{blue}{1.2}}$   & $84.3_{\textcolor{blue}{2.1}}$   & $50.2_{\textcolor{blue}{3.4}}$     & $80.1_{\textcolor{blue}{2.9}}$     & $51.5_{\textcolor{blue}{2.5}}$     & $89.8_{\textcolor{blue}{1.7}}$        & $80.3_{\textcolor{blue}{0.9}}$       & $55.5_{\textcolor{blue}{0.6}}$  \\ \cline{2-13}
& Vanilla ICL   & \multirow{2}{*}{12}          &  $ {70.9}_{\textcolor{blue}{9.6}}$     & ${34.7}_{\textcolor{blue}{6.7}}$ & ${65.2}_{\textcolor{blue}{5.6}}$ & ${59.9}_{\textcolor{blue}{6.7}}$  & $38.3_{\textcolor{blue}{10.2}}$     & ${59.6}_{\textcolor{blue}{8.1}}$  &  ${40.7}_{\textcolor{blue}{7.5}}$    &     ${72.5}_{\textcolor{blue}{11.6}}$    &    ${57.6}_{\textcolor{blue}{9.5}}$    &  ${48.5}_{\textcolor{blue}{5.7}}$\\
& IDAICL  &        &  $90.0_{\textcolor{blue}{2.8}}$     & $41.1_{\textcolor{blue}{1.3}}$      &   $83.4_{\textcolor{blue}{2.3}}$ &  ${85.6}_{\textcolor{blue}{2.4}}$  &   $51.6_{\textcolor{blue}{2.9}}$   &   $80.5_{\textcolor{blue}{2.5}}$   &  $51.8_{\textcolor{blue}{3.6}}$    &      $90.5_{\textcolor{blue}{2.7}}$   &    $81.1_{\textcolor{blue}{3.0}}$    & ${55.7}_{\textcolor{blue}{2.1}}$    \\ \thickcline
& MetaICL    & \multirow{2}{*}{12}     &  ${79.1}_{\textcolor{blue}{7.0}}$     &   $38.6_{\textcolor{blue}{3.7}}$    & $76.4_{\textcolor{blue}{6.3}}$   &  $75.3_{\textcolor{blue}{4.5}}$  &  $50.5_{\textcolor{blue}{7.1}}$    & $73.9_{\textcolor{blue}{7.6}}$     &  $46.7_{\textcolor{blue}{6.3}}$    &   $86.8_{\textcolor{blue}{7.8}}$      &  $76.4_{\textcolor{blue}{5.4}}$    & ${53.1}_{\textcolor{blue}{1.6}}$   \\
&+IDAICL   &   &   ${89.6}_{\textcolor{blue}{2.2}}$    & ${\underline{42.9}}_{\textcolor{blue}{2.3}}$      & ${84.2}_{\textcolor{blue}{3.4}}$   &  ${\underline{87.9}}_{\textcolor{blue}{1.1}}$  &  ${\boldsymbol{53.8}}_{\textcolor{blue}{1.2}}$    &  ${\underline{83.4}}_{\textcolor{blue}{3.2}}$    & ${{53.6}}_{\textcolor{blue}{1.3}}$     & ${\underline{91.9}}_{\textcolor{blue}{0.9}}$        &  ${\underline{84.3}}_{\textcolor{blue}{1.4}}$      & ${\underline{57.3}}_{\textcolor{blue}{1.5}}$   \\ \cline{2-13}
& Channel ICL & \multirow{2}{*}{12}        &  ${83.3}_{\textcolor{blue}{5.9}}$     &   $37.5_{\textcolor{blue}{4.6}}$    & ${80.6}_{\textcolor{blue}{4.1}}$   &  ${77.1}_{\textcolor{blue}{5.5}}$  &  $48.9_{\textcolor{blue}{6.7}}$    & $68.2_{\textcolor{blue}{8.3}}$     &  ${43.3}_{\textcolor{blue}{7.2}}$    &   $70.4_{\textcolor{blue}{9.3}}$      &  ${67.9}_{\textcolor{blue}{5.5}}$    & ${53.6}_{\textcolor{blue}{8.9}}$   \\
& +IDAICL   &   &   ${\boldsymbol{91.2}}_{\textcolor{blue}{2.1}}$    & ${40.8}_{\textcolor{blue}{1.5}}$      & ${\underline{86.5}}_{\textcolor{blue}{2.6}}$   &  ${\boldsymbol{88.2}}_{\textcolor{blue}{1.8}}$  &  ${52.4}_{\textcolor{blue}{2.9}}$    &  ${82.3}_{\textcolor{blue}{2.4}}$    & ${50.5}_{\textcolor{blue}{1.8}}$     & ${88.7}_{\textcolor{blue}{1.2}}$        &  ${82.6}_{\textcolor{blue}{0.9}}$      & ${56.5}_{\textcolor{blue}{2.1}}$   \\ 
\cline{2-13}
% SimCSE     &        &       &    &     &      &       &     &          &         &    \\
% IDA+SimCSE &       &       &    &    &      &      &      &         &        &    \\ \hline
% & $k$NN-LM     &  $70.9_{\textcolor{blue}{7.8}}$      & $44.3_{\textcolor{blue}{8.1}}$      & $68.9_{\textcolor{blue}{6.2}}$   &  $63.8_{\textcolor{blue}{7.1}}$   &  $62.8_{\textcolor{blue}{4.9}}$    &  $66.8_{\textcolor{blue}{7.7}}$     &   $58.7_{\textcolor{blue}{6.6}}$  &     $70.8_{\textcolor{blue}{6.8}}$     &  $78.6_{\textcolor{blue}{7.1}}$       & $72.5_{\textcolor{blue}{7.3}}$   \\
% & $k$NN-LM+IDA &   $77.6_{\textcolor{blue}{4.5}}$    & $46.5_{\textcolor{blue}{3.1}}$       & ${82.3}_{\textcolor{blue}{2.8}}$   &  $84.3_{\textcolor{blue}{4.1}}$  & $\underline{66.9}_{\textcolor{blue}{1.4}}$     &  $76.6_{\textcolor{blue}{3.1}}$    &  ${60.9}_{\textcolor{blue}{3.5}}$    &  $81.7_{\textcolor{blue}{2.4}}$       &  ${83.5}_{\textcolor{blue}{2.4}}$      & $\underline{77.6}_{\textcolor{blue}{3.7}}$   \\ \cline{2-12}
% $k$NN-Prompt     &        &       &    &     &      &       &     &          &         &    \\
% IDA+$k$NN-Prompt &       &       &    &    &      &      &      &         &        &    \\ \hline
% GUNDAM     &        &       &    &     &      &       &     &          &         &    \\
% IDA+GUNDAM &       &       &    &    &      &      &      &         &        &    \\ \hline
& EPR       & \multirow{2}{*}{12}       &   ${82.8}_{\textcolor{blue}{2.6}}$    &  $40.6_{\textcolor{blue}{2.1}}$     &   $79.5_{\textcolor{blue}{1.4}}$ &  $74.7_{\textcolor{blue}{2.7}}$  & $50.7_{\textcolor{blue}{2.3}}$     &  ${{83.3}}_{\textcolor{blue}{0.7}}$    &   ${\underline{82.2}}_{\textcolor{blue}{2.4}}$   & $91.5_{\textcolor{blue}{0.8}}$        &   ${{83.2}}_{\textcolor{blue}{1.6}}$     &  $54.8_{\textcolor{blue}{1.9}}$   \\
& +IDAICL    &      &  ${\underline{90.5}}_{\textcolor{blue}{1.5}}$     &  ${\boldsymbol{43.8}}_{\textcolor{blue}{1.0}}$     & ${\boldsymbol{87.4}}_{\textcolor{blue}{0.9}}$   & ${86.5}_{\textcolor{blue}{1.5}}$    & ${\underline{52.9}}_{\textcolor{blue}{1.8}}$     & ${\boldsymbol{85.8}}_{\textcolor{blue}{0.5}}$     & ${\boldsymbol{84.7}}_{\textcolor{blue}{1.1}}$     & ${\boldsymbol{93.5}}_{\textcolor{blue}{2.5}}$        &  ${\boldsymbol{86.4}}_{\textcolor{blue}{2.2}}$      & ${\boldsymbol{57.5}}_{\textcolor{blue}{1.5}}$    \\
 \midrule\midrule
\multirow{6}*{\rotatebox{90}{GPT-Neo}} 

& MetaICL  & \multirow{2}{*}{12}        & ${87.8}_{\textcolor{blue}{6.7}}$      &   $42.5_{\textcolor{blue}{6.1}}$    &  $82.2_{\textcolor{blue}{5.9}}$  & ${80.7}_{\textcolor{blue}{4.8}}$   & $51.5_{\textcolor{blue}{5.3}}$     &  $72.2_{\textcolor{blue}{8.2}}$    &  $54.1_{\textcolor{blue}{6.8}}$    & $84.4_{\textcolor{blue}{5.5}}$        & $74.3_{\textcolor{blue}{8.2}}$       & ${50.3}_{\textcolor{blue}{6.4}}$    \\
& +IDAICL  &    & ${\underline{92.1}}_{\textcolor{blue}{1.1}}$      & ${{44.3}}_{\textcolor{blue}{2.3}}$      & ${\boldsymbol{88.8}}_{\textcolor{blue}{2.1}}$   & ${\boldsymbol{88.1}}_{\textcolor{blue}{1.8}}$   &  ${\boldsymbol{53.2}}_{\textcolor{blue}{1.7}}$    &  ${{84.3}}_{\textcolor{blue}{2.1}}$    & ${{64.3}}_{\textcolor{blue}{1.9}}$     &  ${{94.3}}_{\textcolor{blue}{1.2}}$       &  ${{86.5}}_{\textcolor{blue}{0.9}}$      & ${\boldsymbol{53.4}}_{\textcolor{blue}{2.1}}$   \\ \cline{2-13}
& Channel ICL  & \multirow{2}{*}{12}        & ${83.4}_{\textcolor{blue}{5.4}}$      &   $39.8_{\textcolor{blue}{6.4}}$    &  $79.5_{\textcolor{blue}{5.7}}$  & ${79.4}_{\textcolor{blue}{5.9}}$   & $50.1_{\textcolor{blue}{3.8}}$     &  $70.6_{\textcolor{blue}{8.2}}$    &  $50.8_{\textcolor{blue}{5.1}}$    & $78.3_{\textcolor{blue}{7.1}}$        & $72.5_{\textcolor{blue}{6.9}}$       & ${48.7}_{\textcolor{blue}{4.5}}$    \\
& +IDAICL  &    & ${91.5}_{\textcolor{blue}{2.2}}$      & ${41.6}_{\textcolor{blue}{1.8}}$      & ${85.4}_{\textcolor{blue}{1.9}}$   & ${\underline{87.2}}_{\textcolor{blue}{2.5}}$   &  ${\underline{52.7}}_{\textcolor{blue}{2.2}}$    &  ${83.7}_{\textcolor{blue}{1.4}}$    & ${62.8}_{\textcolor{blue}{0.7}}$     &  ${93.5}_{\textcolor{blue}{3.3}}$       &  ${84.6}_{\textcolor{blue}{3.1}}$      & ${52.0}_{\textcolor{blue}{1.8}}$   \\ \cline{2-13}
& EPR     & \multirow{2}{*}{12}         &   ${88.2}_{\textcolor{blue}{1.6}}$    &  ${\underline{45.7}}_{\textcolor{blue}{2.2}}$     &   ${81.8}_{\textcolor{blue}{1.9}}$ &  ${71.8}_{\textcolor{blue}{2.9}}$  & ${49.9}_{\textcolor{blue}{1.1}}$     &  ${\underline{89.4}}_{\textcolor{blue}{2.4}}$    & ${\underline{92.3}}_{\textcolor{blue}{2.2}}$     & ${\underline{96.1}}_{\textcolor{blue}{1.2}}$        & ${\underline{88.8}}_{\textcolor{blue}{1.1}}$       & $49.4_{\textcolor{blue}{0.7}}$   \\
& +IDAICL    &      &   ${\boldsymbol{93.2}}_{\textcolor{blue}{0.8}}$    &  ${\boldsymbol{47.2}}_{\textcolor{blue}{1.3}}$     & ${\underline{88.5}}_{\textcolor{blue}{1.2}}$   & ${86.6}_{\textcolor{blue}{2.0}}$   & ${52.1}_{\textcolor{blue}{0.4}}$     &  ${\boldsymbol{93.1}}_{\textcolor{blue}{1.2}}$    & ${\boldsymbol{94.4}}_{\textcolor{blue}{2.4}}$     & ${\boldsymbol{97.8}}_{\textcolor{blue}{1.5}}$        &${\boldsymbol{91.2}}_{\textcolor{blue}{0.7}}$        & ${\underline{52.1}}_{\textcolor{blue}{0.5}}$   \\ 

\bottomrule[1.5pt]
\end{tabular}}
\caption{Comparison results of three PLMs.
% , including GPT-2 with 0.8B and 1.5B parameters and GPT-Neo. 
Two numbers indicate the mean accuracy (\%) and standard deviation over different seeds. The best and second-best results per PLM per dataset are highlighted in bold and underlined, respectively. "+IDAICL" means that the current approach is used in conjunction with IDAICL. The results for different numbers of demonstration examples (i.e., $m$ values) using the GPT-Neo model are illustrated in Figure~\ref{table2fig}.
 }
 % \vspace{-0.1in}
\label{tab1}
\end{table*}
% In the previous derivation, Eq.~(\ref{convex}) is calculated using the linearity of expectation.
% % with the 
% % computed using Jensen’s inequality $\mathbb{E}(\log X)\le \log(\mathbb{E}(X))$, taking into account the concave nature of $\log(\cdot)$. 
% As for Eq.~(\ref{zuizhong}), given that $\boldsymbol{\tilde{h}}_{\mathcal{C}}$ is a Gaussian random variable conforming to $\mathcal{N}\left(\lambda({\boldsymbol{h}}_{\mathcal{C}}+\boldsymbol{\mu}), \lambda\Sigma\right)$, we know that $\Delta \boldsymbol{w}_{k,y_j}^{T} \tilde{\boldsymbol{h}}_{\mathcal{C}}$ follows the multivariant normal distribution: $\mathcal{N}(\lambda\Delta \boldsymbol{w}_{k,y_j}^{T} \left({\boldsymbol{h}}_{\mathcal{C}}+\boldsymbol{\mu}\right), \lambda\Delta \boldsymbol{w}_{k,y_j}^{T} \Sigma \Delta \boldsymbol{w}_{k,y_j})$. Then, Eq.~(\ref{zuizhong}) can be derived by utilizing the moment-generating function:
% \begin{equation}
%     \mathbb{E}[e^{{t}{X}}] = e^{{t}{\mu}+\frac{1}{2}{t}^{2}\sigma^{2}}, X \sim \mathcal{N}(\mu,\sigma^{2}).
% \end{equation}

% As $\boldsymbol{h}_{\boldsymbol{x}}$ in Eq.~(\ref{zuizhong}) is indirectly available in ICL,
% we utilize $\boldsymbol{h}_{\tilde{\boldsymbol{x}}}$, which contains information from both the demonstrations and the query, to replace $\boldsymbol{h}_{\boldsymbol{x}}$.
Subsequently, our newly proposed prediction function, referred to as IDA-Softmax, is defined as
% \footnote{IDA-Softmax remains applicable when each answer comprises multiple tokens as the prediction for each token within the answer can be derived in the same manner.} 
% know that an upper bound for $\ell^{\infty}_{y_j}$ is
    \begin{equation}
    \resizebox{0.99\linewidth}{!}{$
    \begin{aligned}
        P_{y_j}^{\text{IDA}}(\boldsymbol{\tilde{x}}) := \sum_{k} M_{k,y_j}N_{k,y_j}e^{\Delta\boldsymbol{w}_{k,y_j}^{T}\boldsymbol{h}_{\tilde{\boldsymbol{x}}}+\Delta b_{k,y_j}}.
        \end{aligned}
        $}
        \label{prop}
        \end{equation}   
% In $\ell_{y_j}^{\text{IDA}}(\boldsymbol{\tilde{x}})$, $\exp({\Delta\boldsymbol{w}_{k,y_j}^{T}\boldsymbol{h}_{\tilde{\boldsymbol{x}}}+\Delta b_{k,y_j}})$ can be considered as the original scoring function for ICL. 
% In the implementation, $\boldsymbol{h}_{\mathcal{C}}$ in $P^{\text{IDA}}_{y_j}$ corresponds to the hidden state of the last block at the final token position in the demonstration $\mathcal{C}$. 

% Consequently, our proposed augmentation strategy boils down to a novel prediction function, which can be easily adopted by various PLMs.
% we can efficiently achieve the objective of demonstration augmentation by directly employing 
Consequently, instead of conducting the augmentation process explicitly, we can directly employ IDA-Softmax, $P_{y_j}^{\text{IDA}}$, for prediction.
IDA-Softmax essentially utilizes two modulating factors 
associated with statistical properties derived from $\mathcal{D}$ to calibrate the sample logits. Previous studies~\cite{min-etal-2022-rethinking,chan2022data} have underscored the pivotal role of knowledge about the input data distribution in predictions made by PLMs. Intuitively, PLMs can better capture the patterns and underlying structures within data, such as the spatial relationships between demonstrations and queries, ultimately enhancing their prediction performance.

% This manner is reasonable as the input distribution is the key driver of end task performance~\cite{}. 
% Considering that $\exp{(\Delta\boldsymbol{w}_{k,y_j}^{T}\boldsymbol{h}_{\tilde{\boldsymbol{x}}}\!+\!\Delta b_{k,y_j})}$ is the original probability, we understand that IDA-Softmax 
% as accomplished by IDA, 

% We argue that these statistics represent certain underlying patterns that assist large PLMs in making more accurate predictions. 
% \begin{equation}
% \resizebox{0.99\linewidth}{!}{$
% \begin{aligned}
%     &\Delta \boldsymbol{w}_{k,v}^{T} \tilde{\boldsymbol{h}}_{\mathcal{C}}+\Delta b_{k,v} \sim \\&\mathcal{N}(\Delta \boldsymbol{w}_{k,v}^{T} \left({\boldsymbol{h}}_{\mathcal{C}}+\lambda\boldsymbol{\mu}\right)+\Delta b_{k,v}, 
%     \lambda\Delta \boldsymbol{w}_{k,v}^{T} \Sigma \Delta \boldsymbol{w}_{k,v}).
% \end{aligned}$}
% \end{equation}
% Detailed derivation for Eq.~(\ref{zuizhong}) is provided in Section~\ref{ana} of the Appendix. In the above, all features are assumed to share the same dimension, which is easily attainable in implementations. 

% , which harnesses two modulating factors, namely $M_{k,y_j}$ and $N_{k,y_j}$, to calibrate the original scoring function. In real applications, $\boldsymbol{h}_{\boldsymbol{x}}$ is directly replaced by the output hidden state at the last token position, i.e., $\boldsymbol{h}_{\tilde{\boldsymbol{x}}}$.  

Furthermore, to mitigate the imbalance among different answer types in demonstrations~\cite{holtzman-etal-2021-surface,zhao2021calibrate}, we adopt a post-hoc adjustment approach inspired by Menon et al.~\shortcite{menon2020long}, which adjusts predictions by considering the class proportions within $\mathcal{D}$. Thus, the prediction for answer $y_j$ is computed as
\begin{equation}
    \tilde{P}_{y_j}^{\text{IDA}}(\tilde{\boldsymbol{x}}) = P_{y_j}^{\text{IDA}}(\tilde{\boldsymbol{x}}) + \tau\log\pi_{y_j},
\end{equation}
where $\tau$ is a positive hyperparameter, and $\pi_{y_j}$ demotes the proportion of answer $y_j$ in $\mathcal{D}$. 
% The value of $\tau$ can be fixed as 2 in real applications. 
In practical applications, the value of $\tau$ can be fixed at 1. This approach compensates for predictions of minor classes. When different answers are uniformly distributed, $\tau\log\pi_{y_j}$ exerts an equal influence on all answer types. Consequently, the final prediction is given by
\begin{equation}
    \hat{y}={\arg \min_{y_{j}\in\mathcal{Y}} \tilde{P}_{y_j}^{\text{IDA}}(\boldsymbol{\tilde{x}})}.
\end{equation}
% We present IDAICL's pseudo code in Algorithm~\ref{alg1}.
\section{Experimental Setup}
\subsection{Models and Datasets}
We evaluated the performance of IDAICL across seven large PLMs, including GPT-2~\cite{radford2019language} (with 0.1B, 0.3B, 0.8B, and 1.5B parameters), GPT-Neo~\cite{Black2021GPTNeoLS} (with 2.7B parameters), and LLaMA~\cite{touvron2023llama} (with 13B and 33B parameters). 
Following previous research~\cite{min-etal-2022-noisy,han2022prototypical,lu-etal-2022-fantastically}, our evaluation encompasses ten text classification datasets. Among these, SST-2~\cite{socher-etal-2013-recursive}, SST-5~\cite{socher-etal-2013-recursive},
MR~\cite{pang-lee-2005-seeing}, CR~\cite{hu2004mining}, and Amazon~\cite{mcauley2013hidden} are five sentiment classification tasks. Subj~\cite{pang-lee-2004-sentimental}, TREC~\cite{voorhees2000building}, DBPedia~\cite{lehmann2015dbpedia}, and AGNews~\cite{zhang2015character} cater to subjectivity, question, ontology, and news classification tasks, respectively. Additionally, CB~\cite{de2019commitmentbank} is utilized for natural language inference.
 Among these datasets, SST-5, Amazon, TREC, and CB are characterized by imbalanced training data. Details of all datasets are provided in Section~\ref{data} of the Appendix.
% Detailed statistical characteristics and instances for all datasets are presented in 

% Comprehensive information regarding all datasets are presented in Section~\ref{data} of the Appendix.
% statistical characteristics and instances 

% Tables~\ref{data} and~\ref{sample} 

% Detailed statistical characteristics and instances for all datasets are presented in Tables~\ref{data} and~\ref{sample} of the Appendix.  
% four distinct sizes of GPT-2~\cite{radford2019language} (with 0.1B, 0.3B, 0.8B, and 1.5B parameters), along with GPT-Neo~\cite{Black2021GPTNeoLS} (with 2.7B parameters). 
% The experiments are conducted using the open-source checkpoints for GPT-2, GPT-Neo, and LLaMA models.
 % These datasets cover varying numbers of classes per task, ranging from 2 to 14.
% Subj~\cite{pang-lee-2004-sentimental}, 
 % TREC~\cite{li-roth-2002-learning}, DBPedia~\cite{auer2007dbpedia}, and AGNews~\cite{zhang2015character} are subjectivity, question, ontology, and news classification tasks, respectively; and CB~\cite{de2019commitmentbank} is used for natural language inference. 
% The examples from each dataset are placed in Table~\ref{sample}.
\begin{table*}[t]
\small
% \vspace{-0.02in}
\centering
\resizebox{1\linewidth}{!}{
\begin{tabular}{l|l|cccccccccc}
\toprule[1.2pt]
\rowcolor{gray!30}
       PLM & Method       & SST-2 & SST-5 & MR &  CR & Subj & TREC & DBPedia & AGNews & CB & Avg. \\ \midrule\midrule

\multirow{5}*{\rotatebox{90}{LLaMA 13B}} 
 & Vanilla ICL             & ${95.6}_{\textcolor{blue}{7.1}}$ & ${29.5}_{\textcolor{blue}{6.2}}$ & ${90.0}_{\textcolor{blue}{5.8}}$  & ${91.4}_{\textcolor{blue}{7.4}}$ & ${72.9}_{\textcolor{blue}{6.9}}$ & ${62.8}_{\textcolor{blue}{9.1}}$ & ${80.9}_{\textcolor{blue}{7.6}}$ & $80.2_{\textcolor{blue}{5.9}}$ & $51.5_{\textcolor{blue}{8.2}}$ & 72.8\\
& ConCa     &   ${\boldsymbol{96.7}}_{\textcolor{blue}{5.4}}$    &   $40.3_{\textcolor{blue}{6.2}}$ &  ${91.7}_{\textcolor{blue}{7.3}}$          &  ${90.8}_{\textcolor{blue}{4.2}}$  & ${79.6}_{\textcolor{blue}{9.1}}$ & ${68.2}_{\textcolor{blue}{5.6}}$ & ${\underline{94.3}}_{\textcolor{blue}{4.1}}$ & ${\underline{85.2}}_{\textcolor{blue}{7.5}}$ &  $46.6_{\textcolor{blue}{5.0}}$ & {77.0}\\ 
& ${\text{P}}$$\scriptsize{\text{RO}}$${\text{C}}$$\scriptsize{\text{A}}$       &    ${95.4}_{\textcolor{blue}{3.8}}$    &   ${\underline{43.4}}_{\textcolor{blue}{5.7}}$ & ${90.3}_{\textcolor{blue}{9.6}}$      &  ${\underline{92.1}}_{\textcolor{blue}{3.1}}$ & ${\underline{84.8}}_{\textcolor{blue}{2.5}}$ & ${69.9}_{\textcolor{blue}{2.1}}$ & ${92.5}_{\textcolor{blue}{4.9}}$ & ${81.6}_{\textcolor{blue}{3.6}}$ & $51.4_{\textcolor{blue}{4.2}}$ & \underline{77.9} \\ 
& D-ConCa         & 
${96.3}_{\textcolor{blue}{3.8}}$     & 
${42.5}_{\textcolor{blue}{4.5}}$   &   ${\underline{92.0}}_{\textcolor{blue}{4.1}}$       &  
${90.5}_{\textcolor{blue}{2.9}}$ &   
${82.9}_{\textcolor{blue}{4.5}}$ &  ${\underline{73.7}}_{\textcolor{blue}{3.9}}$&  $87.4_{\textcolor{blue}{7.2}}$ & $82.5_{\textcolor{blue}{3.3}}$ & ${\underline{52.2}}_{\textcolor{blue}{4.1}}$ & {77.8}\\
% $k$NN-Prompt     & ${{83.5}}_{\textcolor{blue}{3.3}}$       &    $\underline{40.9}_{\textcolor{blue}{3.2}}$   &   ${{80.5}}_{\textcolor{blue}{3.5}}$ &  $\underline{{84.9}}_{\textcolor{blue}{2.8}}$      & $\underline{54.5}_{\textcolor{blue}{4.4}}$   \\

% SimCSE     &       &       &    &    &      &      &      &         &        &    \\
% IDA+SimCSE &       &       &    &    &      &      &      &         &        &     \\ \hline

 & IDAICL        & ${\boldsymbol{96.7}}_{\textcolor{blue}{2.5}}$      &   ${\boldsymbol{47.1}}_{\textcolor{blue}{1.1}}$ &  ${\boldsymbol{93.0}}_{\textcolor{blue}{1.9}}$  &      ${\boldsymbol{93.3}}_{\textcolor{blue}{0.8}}$ & ${\boldsymbol{87.8}}_{\textcolor{blue}{2.3}}$
& ${\boldsymbol{76.0}}_{\textcolor{blue}{2.6}}$
& ${\boldsymbol{94.9}}_{\textcolor{blue}{1.0}}$ & ${\boldsymbol{87.7}}_{\textcolor{blue}{2.4}}$ & ${\boldsymbol{59.4}}_{\textcolor{blue}{1.9}}$ & $\boldsymbol{81.8}$\\
\midrule\midrule
\multirow{5}*{\rotatebox{90}{LLaMA 33B}} &
Vanilla ICL             & ${95.5}_{\textcolor{blue}{7.2}}$ & ${29.4}_{\textcolor{blue}{5.6}}$ & ${91.7}_{\textcolor{blue}{5.4}}$  & ${\underline{91.5}}_{\textcolor{blue}{8.1}}$ & ${85.1}_{\textcolor{blue}{6.0}}$ & ${70.9}_{\textcolor{blue}{4.4}}$ & ${86.6}_{\textcolor{blue}{4.5}}$ & $76.2_{\textcolor{blue}{6.1}}$ & ${{59.2}}_{\textcolor{blue}{5.3}}$ & 76.2\\ 
& ConCa     &   ${\underline{95.9}}_{\textcolor{blue}{6.5}}$    &   ${39.1}_{\textcolor{blue}{4.4}}$ &  $90.3_{\textcolor{blue}{7.2}}$          &  ${91.2}_{\textcolor{blue}{3.6}}$  & ${74.6}_{\textcolor{blue}{5.7}}$ & ${76.7}_{\textcolor{blue}{6.2}}$ & ${\underline{92.4}}_{\textcolor{blue}{3.9}}$ & $87.3_{\textcolor{blue}{5.7}}$ & ${{57.9}}_{\textcolor{blue}{6.0}}$ & 78.4 \\ & ${\text{P}}$$\scriptsize{\text{RO}}$${\text{C}}$$\scriptsize{\text{A}}$       &    ${95.5}_{\textcolor{blue}{4.2}}$    &   ${{39.2}}_{\textcolor{blue}{6.3}}$ & ${\underline{92.4}}_{\textcolor{blue}{4.1}}$      &  ${91.3}_{\textcolor{blue}{3.5}}$ & ${\underline{88.3}}_{\textcolor{blue}{2.2}}$ & ${64.7}_{\textcolor{blue}{3.8}}$ & $86.9_{\textcolor{blue}{5.1}}$  & ${85.8}_{\textcolor{blue}{7.1}}$ & ${\underline{59.9}}_{\textcolor{blue}{3.8}}$ & {78.2} \\ 
& D-ConCa         & 
${95.4}_{\textcolor{blue}{3.8}}$     & 
${\underline{40.7}}_{\textcolor{blue}{4.5}}$   &   ${92.1}_{\textcolor{blue}{4.2}}$       &  
${91.0}_{\textcolor{blue}{2.9}}$ &   
${76.4}_{\textcolor{blue}{3.6}}$ &  ${\boldsymbol{80.2}}_{\textcolor{blue}{2.1}}$&  $87.6_{\textcolor{blue}{4.2}}$ & ${\underline{87.7}}_{\textcolor{blue}{4.3}}$ & $56.5_{\textcolor{blue}{3.4}}$ & \underline{78.6}\\ 

& IDAICL        & ${\boldsymbol{96.5}}_{\textcolor{blue}{1.1}}$      &   ${\boldsymbol{46.8}}_{\textcolor{blue}{2.4}}$ &  ${\boldsymbol{93.6}}_{\textcolor{blue}{1.3}}$  &      ${\boldsymbol{92.3}}_{\textcolor{blue}{3.3}}$ & ${\boldsymbol{89.3}}_{\textcolor{blue}{2.4}}$
& ${\underline{79.1}}_{\textcolor{blue}{1.5}}$
& ${\boldsymbol{95.6}}_{\textcolor{blue}{2.3}}$ & ${\boldsymbol{88.4}}_{\textcolor{blue}{1.9}}$ & ${\boldsymbol{64.6}}_{\textcolor{blue}{2.8}}$ & $\boldsymbol{82.9}$ \\
% \bottomrule[1.2pt]
\bottomrule[1.2pt]
\end{tabular}}
% }
% \vspace{-0.05in}
\caption{Comparison results of Macro-F1 for the LLaMA model with 13B and 33B parameters, setting $m$ to 4. 
% IDAICL generally surpasses other prediction calibration approaches.
}
\label{tab213b}
\end{table*}
\begin{figure*}[t] 
\centering
% \vspace{-0.02in}
% \includegraphics[width=1\textwidth]{biase928xiawuxiawu.pdf}
% \includegraphics[width=1\textwidth]{fig310201.pdf}
\includegraphics[width=1\textwidth]{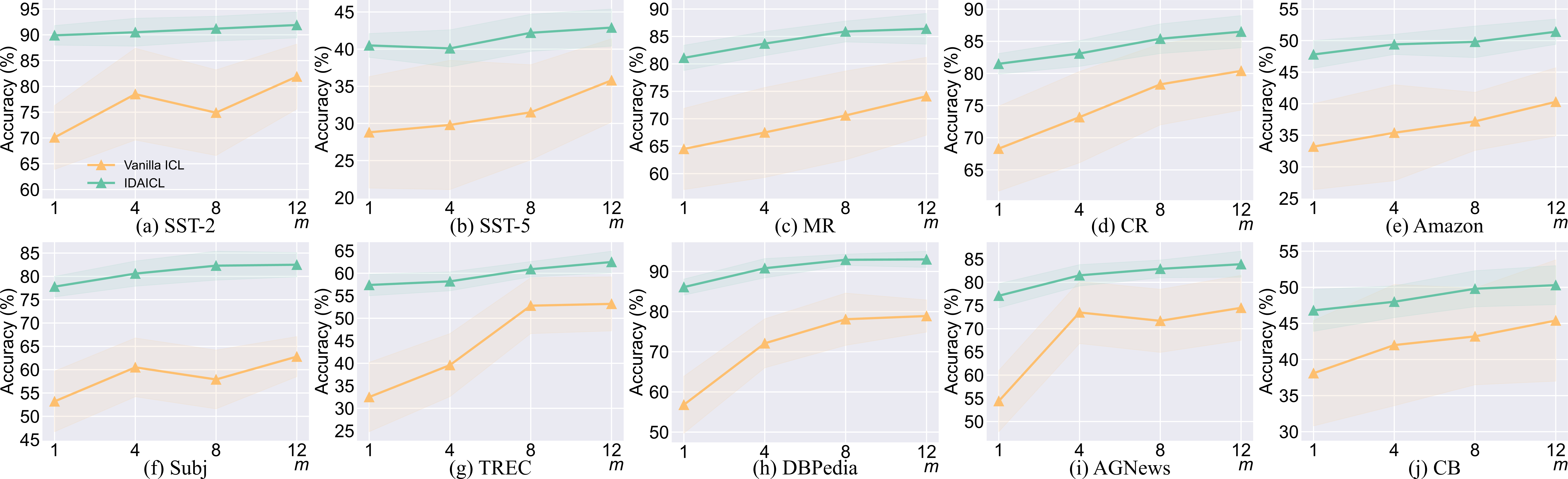}
% \vspace{-0.24in}
\caption{Comparison results between Vanilla ICL and IDAICL across different values of $m$ on the GPT-Neo model. IDAICL significantly outperforms Vanilla ICL, particularly when the number of demonstration examples is small.
% IADICL significantly outper
% The performance of IDAICL remains relatively stable across different numbers of demonstration examples.
% IDAICL demonstrates consistent performance compared to Vanilla ICL across various numbers of demonstration examples.
}
\label{table2fig}
% \vspace{-0.05in}
% \vspace{-0.16in}
\end{figure*}

\subsection{Compared Baselines}
Besides Vanilla ICL, we compared and integrated IDAICL with three popular ICL algorithms, focusing on learning process design and demonstration retrieval. These include MetaICL~\cite{min-etal-2022-metaicl}, Channel ICL~\cite{min-etal-2022-noisy}, and Efficient Prompt Retrieval (EPR)~\cite{rubin-etal-2022-learning}. Moreover, we compared IDAICL with other advanced prediction calibration methods: Contextual Calibration (ConCa)~\cite{zhao2021calibrate}, Prototypical Calibration ($\text{P}$$\small{\text{RO}}$$\text{C}$$\small{\text{A}}$)~\cite{han2022prototypical}, and Domain-Context Calibration (D-ConCa)~\cite{fei-etal-2023-mitigating}. Introductions to all compared methods and comprehensive experimental settings are presented in Sections~\ref{method} and~\ref{setting} of the Appendix.
\begin{figure*}[t] 
\centering
% \vspace{-0.02in}
% \includegraphics[width=1\textwidth]{biase928xiawuxiawu.pdf}
% \includegraphics[width=1\textwidth]{fig310201.pdf}
\includegraphics[width=1\textwidth]{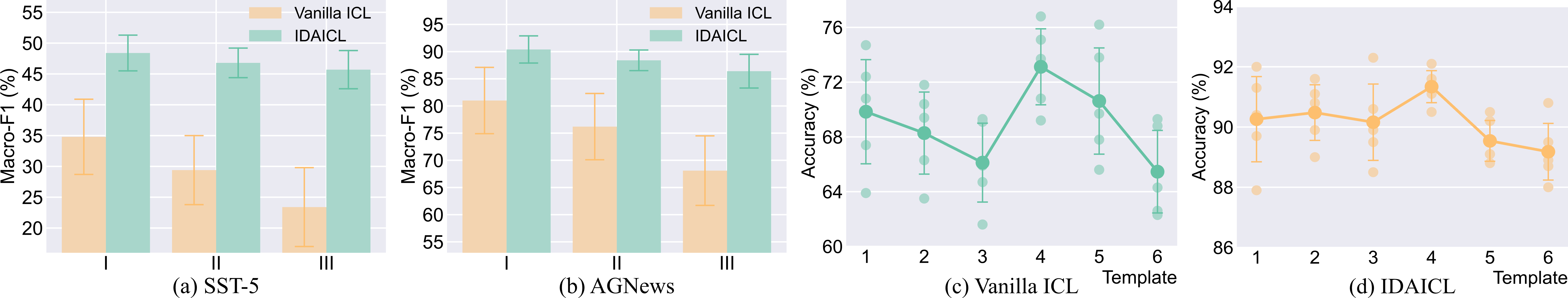}
% \vspace{-0.05in}
\caption{(a) and (b): Macro-F1 of SST-5 and AGNews datasets using the LLaMA model with 33B parameters under three demonstration selection settings, setting $m$ to 4.
(c) and (d): Accuracy of Vanilla ICL and IDAICL on the SST-2 dataset using the GPT-2 model with 1.5B parameters across six templates, setting $m$ to 12. 
IDAICL demonstrates greater robustness across various demonstration examples and templates compared to Vanilla ICL.
% The value of $m$ is set to 12.
}
\label{fig1}
% \vspace{-0.04in}
% \vspace{-0.16in}
\end{figure*}

\section{Experimental Results}

\subsection{Main Results}
% To verify the seamless integration of IDAICL with other ICL approaches to enhance performance, we compare and integrate IDAICL with three popular ICL algorithms: MetaICL~\cite{min-etal-2022-metaicl}, Channel ICL~\cite{min-etal-2022-noisy}, and Efficient Prompt Retrieval (EPR)~\cite{rubin-etal-2022-learning}. Detailed introductions to all compared methods are available in Section~\ref{method} of the Appendix. 
% Additionally, as depicted in Figure~\ref{table2fig}, it exhibits an average improvement of 13.6\% over Vanilla ICL on the GPT-Neo model. 

Table~\ref{tab1} displays the comparison results between IDAICL and four ICL baselines (Vanilla ICL, MetaICL, Channel ICL, and EPR) across GPT-2 models (with 0.8B and 1.5B parameters) and the GPT-Neo model. These results lead to three main findings. Firstly, IDAICL consistently exhibits high effectiveness across various model sizes and datasets, highlighting its strong generalization capacity, even under scenarios involving imbalanced training data. Compared to Vanilla ICL, IDAICL outperforms by an average of 17.7\% and 18.4\% across diverse datasets and $m$ values for GPT-2 with 0.8B and 1.5B parameters, respectively.
Secondly, in comparison to other ICL baselines like Channel ICL, MetaICL, and EPR, the integration of IDAICL consistently delivers notable performance improvements, emphasizing the efficacy of enhancing demonstrations for refined predictions. The inclusion of IDAICL led to an average performance boost of 7.3\% for MetaICL and 8.2\% for Channel ICL. Lastly, IDAICL notably enhances worst-case accuracy and diminishes performance variance across different seeds, showcasing its ability to improve prediction stability. Additional results on LLaMA and smaller GPT-2 models are available in Tables~\ref{llamajiben} and~\ref{gptsmall} of the Appendix.

% We evaluated the effectiveness of IDAICL across all datasets and PLMs. Tables~\ref{tab1} and~\ref{tab213b} present the comparative results for the GPT-2 (with 0.8B and 1.5B parameters), GPT-Neo, and LLaMA models. Others are presented in Section~\ref{comp} of the Appendix.
% Table~\ref{gptsmall}
% Additional results are placed in Section~\ref{comp} 
% of the Appendix. 

\subsection{Comparison with Calibration Methods}
We compared IDAICL with three advanced prediction calibration methods (ConCa, $\text{P}$$\small{\text{RO}}$$\text{C}$$\small{\text{A}}$, and D-ConCa) across three PLMs: GPT-2, GPT-Neo, and LLaMA. Table~\ref{tab213b} presents the comparison results for the LLaMA models, where IDAICL consistently achieves state-of-the-art performance, except for TREC using the LLaMA model with 33B parameters. These findings suggest that IDAICL which leverages statistical information derived from the input data distribution for prediction calibration, generally outperforms methods relying on estimated biases for correction. Further comparison results can be found in Table~\ref{tabcalibrating} of the Appendix.

% We proceed to compare IDAICL with other advanced prediction calibration methods (ConCa, $\text{P}$$\small{\text{RO}}$$\text{C}$$\small{\text{A}}$, and D-ConCa) on three PLMs. Table~\ref{tab213b} presents the comparison results for the LLaMA model. Compared to methods that use estimated biases for prediction correction, IDAICL, which relies on statistical information derived from input distribution for prediction calibration, achieves superior performance. Further comparison results can be found in Table~\ref{tabcalibrating} of the Appendix.
% for GPT-2 with 1.5B parameters can be found in Table~\ref{tabcalibrating} of the Appendix. 
% Furthermore, IDAICL proves effective in addressing challenges associated with class-imbalanced training data, as evidenced by results of SST-5, Amazon, TREC, and CB datasets. More detailed analysis can be found in Section~\ref{imbalance}.

% \begin{figure}[t] 
% \centering
% % \vspace{-0.02in}
% \includegraphics[width=0.48\textwidth]{fig310111.pdf}
% % \vspace{-0.05in}
% \caption{Accuracy on SST-2 and MR datasets for different values of $m$.}
% \label{mvary}
% % \vspace{-0.02in}
% % \vspace{-0.16in}
% \end{figure}

% \begin{figure}[bp] 
% \centering
% \vspace{-0.05in}
% % \includegraphics[width=0.48\textwidth]{temp1019111.pdf}
% \includesvg[width=0.48\textwidth]{fig4112.svg}
% % \vspace{-0.12in}
% \caption{Accuracy across six templates on SST-2.}
% \label{template}
% % \vspace{-0.02in}
% % \vspace{-0.16in}
% \end{figure}

\subsection{Stability Analysis}
Previous studies~\cite{zhao2021calibrate,sorensen-etal-2022-information,min-etal-2022-noisy,zhang-etal-2022-active} have highlighted the considerable variability in ICL's performance. In this section, we verified that IDAICL can effectively enhance performance stability across diverse scenarios. 
% across varying quantities of demonstrations, different demonstration selections and permutations, various templates, and imbalanced label distributions.
\paragraph{Varying numbers of demonstrations}
We have presented the results across different numbers of demonstrations in Table~\ref{tab1}. For a clearer depiction, the outcomes regarding GPT-Neo are illustrated in Figure~\ref{table2fig}. As the number of demonstration examples (represented by $m$) increases, both Vanilla ICL and IDAICL exhibit improved performance, emphasizing the importance of comprehensive statistical properties of the input data for IDAICL's effectiveness. Notably, IDAICL significantly enhances performance stability across various numbers of demonstrations and consistently outperforms Vanilla ICL. The performance improvement
is particularly pronounced when $m$ takes on smaller values, indicating the efficacy of IDAICL in enriching the available knowledge for PLMs. 
% without extending the input length.

% which is attributed to the fact that very limited knowledge can be transmitted from demonstrations to PLMs~\cite{li2023context}. Nevertheless, IDAICL effectively enriches the available knowledge for PLMs without extending the input length.

% With the expansion of the demonstration set, the performance of IDAICL exhibits an upward trend, 
% the performance of IDAICL improves, 
% underscoring the significance of a comprehensive 
% % and precise 
% input distribution for augmenting demonstrations. 
% When the number of demonstrations for each test sample reaches 12, the model's performance stabilizes.
% Nevertheless, once the set reaches a certain size, the model's performance stabilizes. 
% of IDAICL relative to Vanilla ICL 

% the limited knowledge transfer from the training data to PLMs when $m$ is small, whereas 

\paragraph{Varying demonstrations}
 To confirm that augmenting demonstrations can enhance the robustness of the ICL strategy across various demonstrations, we investigated three distinct demonstration selection settings. 
% on the SST-5 and AGNews datasets.
Setting I: Training samples most similar to the test sample are chosen. Setting II: Samples are randomly selected from the training data. Setting III: Training samples exhibiting the greatest dissimilarity from the test sample are selected. 
As shown in Figures~\ref{fig1}(a) and (b), IDAICL significantly outperforms Vanilla ICL and demonstrates greater robustness across the three selection settings. Additionally, our discoveries suggest that selecting demonstrations that are more similar to the test samples leads to better performance than exclusively selecting dissimilar ones, which aligns with the findings obtained by Wang et al.~\shortcite{wang-etal-2022-training}.

\begin{figure*}[t] 
\centering
% \vspace{-0.05in}
% \includegraphics[width=1\textwidth]{biase928xiawuxiawu.pdf}
\includegraphics[width=1\textwidth]{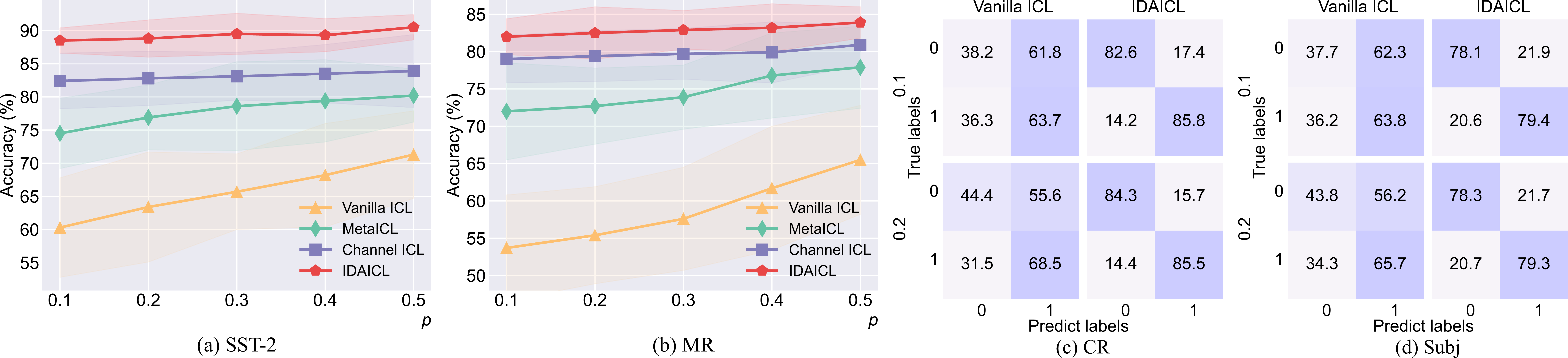}
% \includegraphics[width=1\textwidth]{imbchangshi.pdf}
% \vspace{-0.02in}
% at varying degrees of class imbalance within the demonstrations
% showcasing two distinct levels of imbalance.
% with the proportions of the negative class in demonstrations varied from 0.1 to 0.5. 
% , 
\caption{(a) and (b): Accuracy comparison of the SST-2 and MR datasets, where the proportions of the negative class in demonstrations (denoted as $p$) are varied from 0.1 to 0.5. 
(c) and (d): Confusion matrices for the CR and Subj datasets, representing scenarios where the proportions of one category in demonstrations are set to 0.1 and 0.2. The analysis is conducted using the GPT-2 model with 1.5B parameters, with $m$ setting to 12. 
% The GPT-2 model with 1.5B parameters is employed, setting $m$ to 12. 
IDAICL demonstrates greater robustness in handling imbalanced class distributions within demonstrations.}
\label{biasedata}
% \vspace{-0.05in}
% \vspace{-0.16in}
\end{figure*}

\begin{figure}[t] 
\centering
% \vspace{-0.05in}
% \includegraphics[width=1\textwidth]{biase928xiawuxiawu.pdf}
\includegraphics[width=0.48\textwidth]{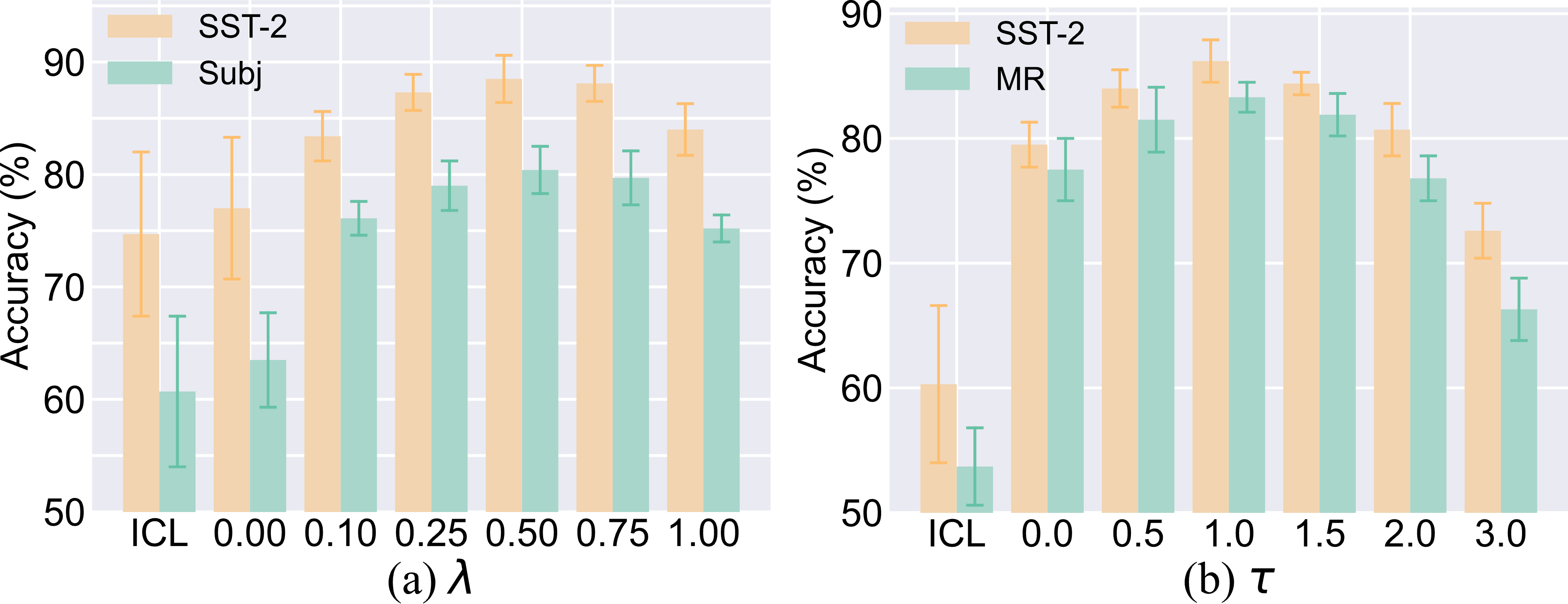}
% \includegraphics[width=1\textwidth]{imbchangshi.pdf}
% \vspace{-0.25in}
% at varying degrees of class imbalance within the demonstrations
% showcasing two distinct levels of imbalance.
% with the proportions of the negative class in demonstrations varied from 0.1 to 0.5. 
% , 
\caption{Accuracy across different $\lambda$ and $\tau$ values, using GPT-2 with 0.8B parameters, setting $m$ to 12. $\lambda\!=\!0$ and $\tau\!=\!0$ signify that the two modulating factors and the class proportion term are not utilized, respectively.
}
\label{sense1212}
% \vspace{-0.1in}
% \vspace{-0.16in}
\end{figure}

% \begin{figure}[t] 
% % \vspace{-0.03in}
% \centering
% % \vspace{-0.02in}
% \includegraphics[width=0.48\textwidth]{fig41012.pdf}
% % \vspace{-0.05in}
% \caption{Accuracy on SST-2 and SST-5 datasets for three different demonstration selection settings.}
% \label{figbijiao}
% % \vspace{-0.02in}
% % \vspace{-0.16in}
% \end{figure}

\paragraph{Varying templates}

% The performance of ICL on GPT-like models has been observed to be sensitive to hand-crafted templates~\cite{sorensen-etal-2022-information,zhao2021calibrate}. 
To assess the performance of IDAICL across various templates, we employed fifteen templates on the SST-2 dataset following those outlined by Zhao et al.~\shortcite{zhao2021calibrate}. The templates are elaborated in Table~\ref{formulate} of the Appendix.
Figures~\ref{fig1}(c) and (d) display the performance of Vanilla ICL and IDAICL across six templates. Some templates achieve higher average performance than others. Nevertheless, IDAICL consistently enhances both average and worst-case accuracy, simultaneously reducing performance variance across different templates. The complete results are available in Figure~\ref{tempfulu} of the Appendix. 
% The templates are detailed in Table~\ref{formulate} of the Appendix. 
% The GPT-2 model with 1.5B parameters is utilized, with $m$ setting to 12. 
% The primary reason behind the effectiveness of IDAICL is its ability to convey a substantial amount of knowledge from the training data to PLMs.

\paragraph{Impact of imbalance in labels}
\label{imbalance}
% Previous algorithms on ICL generally depend on clean and balanced data sources~\cite{wang2023investigating}. However, in real-world scenarios, meeting these conditions can be challenging and costly. We assess the performance of IDAICL in the presence of imbalanced and noisy training data. The experiments are conducted using GPT-2 with 1.5B parameters setting $m$ to 16.
% \subsubsection{Imbalanced Training Data}

% Previous research~\cite{zhao2021calibrate,min-etal-2022-noisy} has confirmed that class imbalance in demonstrations detrimentally affects ICL's performance. 
% We evaluated the capacity of IDAICL for addressing label bias within demonstrations. 
Figures~\ref{biasedata}(a) and (b) depict comparison results among Vanilla ICL, MetaICL, Channel ICL, and IDAICL across different degrees of imbalances.
% on the SST-2 and MR datasets.
% , where the proportions of the negative class in demonstrations are varied from 0.1 to 0.5. 
It is evident that the performance of Vanilla ICL is sensitive to class imbalance, while that of IDAICL and Channel ICL exhibit robustness to the imbalance. Moreover, notable performance improvements are observed with higher levels of imbalance. Additionally, Figures~\ref{biasedata}(c) and (d) illustrate the confusion matrices for CR and Subj datasets, with the proportion of one category (i.e., "Negative" and "Subjective") in demonstrations 
% (i.e., "Negative" in CR and "Subjective" in Subj) 
setting to 0.1 and 0.2. IDAICL significantly improves the accuracy of the underrepresented classes when compared to Vanilla ICL, thereby contributing to enhanced fairness among classes. In the subsequent section, we demonstrate that the strong performance of IDAICL in handling imbalanced label distributions stems from both the statistical properties and the class proportion term.

\begin{table}[bp]
\small
% \vspace{-0.05in}
\centering
\begin{tabular}{l|cccc}
\toprule[1.2pt]
\rowcolor{gray!30}
Dataset & 0-shot   & 1-shot   & 4-shot   & IDAICL   \\  \midrule\midrule
SST-2                 & $63.2$    & $61.3_{\textcolor{blue}{9.4}}$    & $57.6_{\textcolor{blue}{7.1}}$  & 
$\boldsymbol{76.3}$  \\
SST-5                 & $25.0$  & $27.3_{\textcolor{blue}{7.9}}$  & $30.4_{\textcolor{blue}{6.3}}$  & $\boldsymbol{33.5}$  \\ 
MR                 & $58.9$  & $54.3_{\textcolor{blue}{6.8}}$  & $59.3_{\textcolor{blue}{6.5}}$  & $\boldsymbol{71.2}$  \\ 
Subj                & $48.9$  & $47.1_{\textcolor{blue}{8.3}}$  & $57.6_{\textcolor{blue}{5.4}}$  & $\boldsymbol{67.3}$  \\ \bottomrule[1.2pt]
\end{tabular}
\caption{Accuracy comparison between Vanilla ICL and IDAICL based solely on statistical properties, using the GPT-2 model with 0.8B parameters. 
% The experiments were conducted using the GPT-2 model with 0.8B parameters.
}
\label{table4}
% \vspace{-0.05in}
\end{table}

% \begin{table}[t]
% \centering
% \small
% \vspace{-0.08in}
% \begin{tabular}{l|c|c}
% \toprule[1.2pt]
% Model                  & Params & Additional cost \\ \midrule\midrule
% \multirow{3}{*}{GPT-2} & 0.3B   & $6.2\%_{\textcolor{blue}{1.6}}$                \\
%                        & 0.8B   & $6.0\%_{\textcolor{blue}{2.2}}$                \\
%                        & 1.5B   &  $5.5\%_{\textcolor{blue}{1.3}}$               \\ \hline
% GPT-Neo                  & 2.7B   &  $5.7\%_{\textcolor{blue}{2.0}}$              \\ \bottomrule[1.2pt]
% \end{tabular}
% \caption{Additional time cost incurred by IDAICL. 
% % compared with Vanilla ICL.
% }
% \label{time}
% \vspace{-0.1in}
% \end{table}

% \subsection{Prediction Efficiency}
% We have calculated the average additional inference time incurred by IDAICL in comparison to Vanilla ICL across all datasets utilizing an NVIDIA A100 80G GPU. From the results reported in Table~\ref{time}, IDAICL only imposes a negligible computational overhead when compared to Vanilla ICL, revealing the efficiency of our proposed IDAICL.
% proves to be efficient than using explicit demonstration augmentation.

\subsection{Sensitivity and Ablation Studies}
\label{secsense11}

We conducted ablation studies on IDAICL to investigate the influence of the two modulating factors and the class proportion term. The parameters $\lambda$ and $\tau$ govern the augmentation strength and the impact of the class proportion term, respectively. In Figure~\ref{sense1212}(a), a significant performance drop is observed when predictions are not calibrated using statistical properties derived from the demonstrations. Additionally, optimal performance is achieved when $\lambda$ equals 0.5. 

Figure~\ref{sense1212}(b) showcases the accuracy of SST-2 and MR datasets with the negative class proportion in demonstrations setting to 0.1. Results indicate that solely leveraging statistical properties (i.e., $\tau$ equals 0) enhances performance under imbalanced demonstrations, with further improvements observed upon the inclusion of the class proportion term. Additionally, optimal performance is attained when $\tau$ equals 1. Consequently, we recommend setting $\lambda$ to 0.5 and $\tau$ to 1 for practical applications. 
More results are presented in Appendix~\ref{abafulusection}.
% when handling imbalanced demonstrations with 

\subsection{Further Discussion}
To further investigate the effect of statistical properties within demonstrations on model performance, we exclusively employed queries along with statistical information for inference, excluding the inclusion of demonstrations for each test sample. 
These statistics were estimated using deep features of all training samples. As shown in Table~\ref{table4}, IDAICL relying solely on statistical properties distinctly outperforms Vanilla ICL across scenarios with zero, one, and even four demonstrations. This emphasizes the crucial role of prior statistics obtained from training data in PLMs' predictions. This phenomenon is understandable as statistical properties inherently encompass richer global information compared to individual demonstrations.

\section{Conclusion}

This study introduces IDAICL, a novel ICL approach designed to enhance demonstrations by utilizing semantic directions sampled from the deep feature distribution of demonstration examples. Our augmentation strategy enriches the knowledge available to PLMs without extending the context length. A new prediction function is then theoretically established 
% To enhance efficiency, 
considering the number of augmented pieces approaching infinity.
% , we derive a new prediction function, termed IDA-Softmax.
% , which employs two modulating factors to calibrate the original sample logits. 
This eliminates the need for explicit augmentation and allows for direct utilization of this derived function for predictions. Our extensive experiments, spanning various tasks and PLMs, demonstrate that IDAICL significantly enhances both prediction accuracy and stability when compared to other ICL baselines. 
% associated with the statistical properties of the input distribution 
% can achieve considerably better performance than other ICL baselines.

% improves the generalization and robustness of PLMs' predictions.
% Moreover, it can seamlessly integrate with other ICL methods, thereby offering additional performance enhancements.

\section*{Limitations}
% parameter
% math decomposition
% OOD problem limited o training data
% We leave extensions to black-box PLMs for future work.
While IDAICL proves to be competitive in few-shot learning, there are limitations that open up avenues for future research. First, due to the necessity of accessing the parameters of the final fully connected layer in PLMs, IDAICL is exclusively suitable for open-source models. Future research is expected to develop alternative augmentation strategies tailored for black-box PLMs. 
% \textcolor{red}{Second, our method involves two manually set parameters. Nevertheless, our experiments have demonstrated consistent success across different datasets and models with fixed recommended values.}
Second, our evaluation of IDAICL focused on seven PLMs and ten text classification tasks. 
% Subsequent research could delve into the applicability of IDAICL across other PLMs and non-classification tasks. 
We defer further explorations involving other PLMs and non-classification tasks for future work. 
Additionally, IDAICL relies on a small set of demonstrations to estimate the feature mean and covariance matrix. If such a collection is unavailable or extremely scarce, IDAICL may need to be used in conjunction with demonstration generation methods.
% IDAICL may need to be supplemented with some demonstration generation methods.

Other avenues for future work involve exploring more effective augmentation distributions. This entails exploring finer-grained distributions, such as category-level or sample-level distributions, to emphasize the unique characteristics of individual categories or samples, and extending these distributions beyond the constraints of training data. Furthermore, given the effectiveness of data augmentation in model training, future research could explore the utilization of our derived prediction function in both the training and fine-tuning phases of large PLMs.

\bibliography{anthology,custom}
\bibliographystyle{acl_natbib}

% \newpage
\appendix

\begin{table*}[t]
\small
\centering
% \resizebox{1\linewidth}{!}{
\begin{tabular}{l|lccc}
\toprule[1.2pt]
\rowcolor{gray!30}
Dataset & Task                                  & Avg. length & Classes &  Balanced\\ \midrule\midrule
SST-2~\cite{socher-etal-2013-recursive}   & Sentiment analysis            & 12.4        & 2   & Yes    \\
SST-5~\cite{socher-etal-2013-recursive}   & Sentiment analysis            & 23.1        & 5   & No    \\
MR~\cite{pang-lee-2005-seeing}      & Sentiment analysis            & 25.7        & 2   & Yes    \\
CR~\cite{hu2004mining}      & Sentiment analysis      & 22.1        & 2  & Yes     \\
Amazon~\cite{mcauley2013hidden}    & Sentiment analysis             & 78.5         & 5    & No  \\
Subj~\cite{pang-lee-2004-sentimental}    & Subjectivity classification           & 28.9        & 2  & Yes     \\
TREC~\cite{voorhees2000building}    & Question classification & 11.6        & 6  & No      \\
DBPedia~\cite{lehmann2015dbpedia} & Ontology classification               & 65.5        & 14   & Yes   \\
AGNews~\cite{zhang2015character}  & News classification           & 53.8        & 4   & Yes    \\
CB~\cite{de2019commitmentbank}      & Natural language inference            & 69.7/8.4    & 3   & No    \\ \bottomrule[1.2pt]
\end{tabular}
% }
\caption{Statistical information of ten datasets. The average length is calculated based on the GPT-2 sentence-piece length. For tasks involving sentence pairs, we provide the average length for each individual sentence.}
\label{tabledata}
\end{table*}

\begin{table*}[t]
\centering
\small
% \resizebox{.98\linewidth}{!}{
\begin{tabular}{l|m{8.6cm}<{\centering}|m{4.5cm}<{\centering}}
\toprule[1.2pt]
\rowcolor{gray!30}
Dataset & Instances & Label names \\
\midrule\midrule
SST-2       & \makecell[l{p{8.5cm}}]{1. This movie is amazing! (Label = "Positive")\\2. Horrific movie, don’t see it. (Label = "Negative")}       &   Positive, Negative          \\ \hline
 SST-5       &  \makecell[l{p{8.5cm}}]{1. A pretensions – and disposable story — sink the movie. (Label = "Great")\\2. Apparently reassembled from the cutting-room floor of any given daytime soap. (Label = "Terrible")}      &            Terrible, Bad, Okay, Good, Great \\ \hline
 MR &  \makecell[l{p{8.5cm}}]{1. Lame sweet home leaves no southern stereotype unturned. (Label = "Negative")\\2. Not so much farcical as sour. (Label = "Negative")} &  Negative, Positive \\ \hline
CR &  \makecell[l{p{8.5cm}}]{1. It takes excellent pics and is very easy to use, if you read the manual. (Label = "Negative")\\2. Bluetooth does not work on this phone. (Label = "Negative")} & Negative, Positive \\ \hline
Amazon &  \makecell[l{p{8.5cm}}]{1.  Don't waste your money if you already have 2003... There isn’t one reason to get this update if you already have MS
Money 2003 Deluxe and Business. (Label ="Terrible")\\ 2.  The game was in perfect condition! came before it said it should have by 2 days!! I love the game and I suggest it to
a lot of my friends! (Label ="Great")} &  Terrible, Bad, Okay, Good, Great \\ \hline
Subj & \makecell[l{p{8.5cm}}]{1. This is a story about the warm relationship between a little girl and her father despite the difficult conditions they
have to live in. (Label = "Objective")\\2. Too slow, too boring, and occasionally annoying. (Label = "Subjective")} & Subjective, Objective\\ \hline
TREC & \makecell[l{p{8.5cm}}]{1. When did the neanderthal man live? (Label = "Number")\\ 2. How do you get a broken cork out of a bottle? (Label = "Description")} & Description, Entity, Expression, Human, Location, Number\\ \hline
DBPedia  & \makecell[l{p{8.5cm}}]{1. CMC Aviation is a charter airline based in Nairobi Kenya. (Label = "Company")\\2. Dialectica aemula is a moth of the Gracillariidae family. (Label = "Animal")}  &Company, School, Artist, Athlete, Politics, Transportation, Building, Nature, Village, Animal, Plant, Album, Film, Book\\ \hline
AGNews& \makecell[l{p{8.5cm}}]{1. Walk in park for Yankees Drained by a difficult week, the New York Yankees needed an uplifting victory. (Label = "Sports")\\ 2. NASA Mountain View claims world’s fastest computer. (Label = "Technology")} & World, Sports, Business, Technology\\ \hline
CB & \makecell[l{p{8.5cm}}]{1. It was a complex language. Not written down but handed down. One might say it was peeled down.\\
The language was peeled down.\\ (Label = "True")\\2. ``Do you mind if I use your phone?'' Ronni could see that Guido's brain was whirring.\\Guido's brain was whirring. \\ (Label = "True") }   & True, False, Neither \\ 
% \midrule
\bottomrule[1.2pt]
\end{tabular}
% }
\caption{Examples and label names from all datasets.}
\label{sample}
\end{table*}

\begin{table*}[t]
% \vspace{-0.1in}
\centering
\small
\begin{tabular}{l|l|l}
\toprule[1.2pt]
\rowcolor{gray!30}
Dataset & Template & Label mapping \\
\midrule\midrule
SST-2   &  \makecell[l]{Review: \{Sentence\}\\
Sentiment: \{Label\}}       &   Positive / Negative            \\ \hline
SST-5   &    \makecell[l]{Review: \{Sentence\}\\
Sentiment: \{Label\}}     &     terrible / bad / okay / good / great       \\ \hline
MR      &    \makecell[l]{Review: \{Sentence\}\\
Sentiment: \{Label\}}     &      Positive / Negative         \\ \hline
CR      &    \makecell[l]{Review: \{Sentence\}\\
Sentiment: \{Label\}}     &        Positive / Negative       \\ \hline
Subj    &   \makecell[l]{Input: \{Sentence\}\\
Type: \{Label\}}     &         objective / subjective       \\ \hline
TREC    &   \makecell[l]{Question: \{Sentence\}\\
Type: \{Label\}}     &            description / entity / expression / human / location / number   \\ \hline
Amazon  &    \makecell[l]{Review: \{Sentence\}\\
Sentiment: \{Label\}}     &     terrible / bad / okay / good / great \\ \hline
AGNews  &    \makecell[l]{Input: \{Sentence\}\\
Type: \{Label\}}     &          world / sports / business / technology     \\ \hline
DBPedia & \makecell[l]{Input: \{Sentence\}\\
Type: \{Label\}}        &\makecell[l]{company / school / artist / athlete / politics / transportation\\building / nature / village / animal / plant / album / film / book}         \\ \hline
CB      &   \makecell[l]{Premise: \{Sentence\}\\ Hypothesis:
\{Sentence\}\\Prediction: \{Label\}}     & true / false / neither \\
\bottomrule[1.2pt]
\end{tabular}
\caption{Prompt templates and label mappings for each dataset.}
\label{temp}
\end{table*}

\section{Details of Applied Datasets}
\label{data}

Table~\ref{tabledata} presents comprehensive statistics for all datasets utilized in this study. The information includes task descriptions, average sentence lengths, class counts, and details on class imbalance. Additionally, Table~\ref{sample} provides sample instances and label names for each of the datasets.

\begin{table*}[t]
\small
\centering
\resizebox{1\linewidth}{!}{
\begin{tabular}{l|l|c|cccccccccc}
\toprule[1.5pt]
\rowcolor{gray!30}
             PLM & Method & m  & SST-2 & SST-5 & MR & CR & Subj & TREC & DBPedia & AGNews & CB & Avg.\\ \midrule\midrule
\multirow{4}{*}{{13B}} 
 % & Vanilla ICL & \multirow{2}{*}{1}             &  ${45.6}_{\textcolor{blue}{5.8}}$     &   ${20.9}_{\textcolor{blue}{6.7}}$    &  ${40.2}_{\textcolor{blue}{6.1}}$  &  ${47.3}_{\textcolor{blue}{5.9}}$  & $27.0_{\textcolor{blue}{7.1}}$     & ${37.8}_{\textcolor{blue}{5.2}}$     &   ${26.7}_{\textcolor{blue}{4.4}}$   & ${34.7}_{\textcolor{blue}{5.9}}$        &    ${28.6}_{\textcolor{blue}{9.1}}$    & ${35.2}_{\textcolor{blue}{4.3}}$   \\ 
% & IDAICL &         &  $56.2_{\textcolor{blue}{3.1}}$     & $28.5_{\textcolor{blue}{2.7}}$      & $58.1_{\textcolor{blue}{1.8}}$   & ${58.0}_{\textcolor{blue}{2.5}}$   &   $38.6_{\textcolor{blue}{3.4}}$   & $50.3_{\textcolor{blue}{1.9}}$      &    ${41.6}_{\textcolor{blue}{2.2}}$  &  $66.5_{\textcolor{blue}{1.8}}$       &     $51.2_{\textcolor{blue}{1.9}}$   & $59.0_{\textcolor{blue}{2.1}}$   \\ \cline{2-13}
& Vanilla ICL & \multirow{2}{*}{4}             & ${95.6}_{\textcolor{blue}{7.1}}$ & ${29.5}_{\textcolor{blue}{6.2}}$ & ${90.0}_{\textcolor{blue}{5.8}}$  & ${91.4}_{\textcolor{blue}{7.4}}$ & ${72.9}_{\textcolor{blue}{6.9}}$ & ${62.8}_{\textcolor{blue}{9.1}}$ & ${80.9}_{\textcolor{blue}{7.6}}$ & $80.2_{\textcolor{blue}{5.9}}$ & $51.5_{\textcolor{blue}{8.2}}$ & 72.8 \\ 
& IDAICL &         &  ${{96.7}}_{\textcolor{blue}{2.5}}$      &   ${{47.1}}_{\textcolor{blue}{1.1}}$ &  ${{93.0}}_{\textcolor{blue}{1.9}}$  &      ${{93.3}}_{\textcolor{blue}{0.8}}$ & ${{87.8}}_{\textcolor{blue}{2.3}}$
& ${{76.0}}_{\textcolor{blue}{2.6}}$
& ${{94.9}}_{\textcolor{blue}{1.0}}$ & ${{87.7}}_{\textcolor{blue}{2.4}}$ & ${{59.4}}_{\textcolor{blue}{1.9}}$ & ${81.8}$
   \\ \cline{2-13}

& Vanilla ICL & \multirow{2}{*}{8}             &  ${96.7}_{\textcolor{blue}{7.1}}$     &   ${39.4}_{\textcolor{blue}{5.6}}$    &  ${92.3}_{\textcolor{blue}{6.2}}$  &  ${92.2}_{\textcolor{blue}{4.8}}$  & ${70.8}_{\textcolor{blue}{5.1}}$     &   ${71.2}_{\textcolor{blue}{9.1}}$   & ${83.7}_{\textcolor{blue}{4.2}}$        &    ${79.5}_{\textcolor{blue}{6.3}}$    & ${52.4}_{\textcolor{blue}{3.7}}$ &  75.4 \\ 
& IDAICL &         &  $96.9_{\textcolor{blue}{2.1}}$     & $49.2_{\textcolor{blue}{1.9}}$      & $93.4_{\textcolor{blue}{1.6}}$   & ${92.9}_{\textcolor{blue}{1.9}}$   &   $87.5_{\textcolor{blue}{3.0}}$      &    ${79.9}_{\textcolor{blue}{2.1}}$  &  $93.6_{\textcolor{blue}{0.9}}$       &     $88.0_{\textcolor{blue}{1.7}}$   & $62.4_{\textcolor{blue}{2.5}}$  &  82.6 \\
\midrule\midrule
\multirow{4}{*}{{33B}} 
% & Vanilla ICL & \multirow{2}{*}{1}             &  ${44.6}_{\textcolor{blue}{6.9}}$     &   ${19.8}_{\textcolor{blue}{8.2}}$    &  ${40.1}_{\textcolor{blue}{5.6}}$  &  ${46.0}_{\textcolor{blue}{4.7}}$  & $29.8_{\textcolor{blue}{5.1}}$     & ${48.9}_{\textcolor{blue}{5.3}}$     &   ${25.6}_{\textcolor{blue}{7.1}}$   & ${35.8}_{\textcolor{blue}{8.2}}$        &    ${26.9}_{\textcolor{blue}{4.9}}$    & ${30.6}_{\textcolor{blue}{8.1}}$   \\ 
% & IDAICL &         &  $70.2_{\textcolor{blue}{1.9}}$     & $28.8_{\textcolor{blue}{2.3}}$      & $62.1_{\textcolor{blue}{2.4}}$   & ${59.1}_{\textcolor{blue}{1.7}}$   &   $39.6_{\textcolor{blue}{2.5}}$   & $61.1_{\textcolor{blue}{3.2}}$      &    ${48.7}_{\textcolor{blue}{2.9}}$  &  $67.0_{\textcolor{blue}{3.3}}$       &     $48.1_{\textcolor{blue}{3.5}}$   & $56.9_{\textcolor{blue}{1.9}}$   \\ \cline{2-13}
& Vanilla ICL & \multirow{2}{*}{4}             &  ${95.5}_{\textcolor{blue}{7.2}}$ & ${29.4}_{\textcolor{blue}{5.6}}$ & ${91.7}_{\textcolor{blue}{5.4}}$  & ${{91.5}}_{\textcolor{blue}{8.1}}$ & ${85.1}_{\textcolor{blue}{6.0}}$ & ${70.9}_{\textcolor{blue}{4.4}}$ & ${86.6}_{\textcolor{blue}{4.5}}$ & $76.2_{\textcolor{blue}{6.1}}$ & ${{59.2}}_{\textcolor{blue}{5.3}}$ & 76.2  \\ 
& IDAICL &     &     ${{96.5}}_{\textcolor{blue}{1.1}}$      &   ${{46.8}}_{\textcolor{blue}{2.4}}$ &  ${{93.6}}_{\textcolor{blue}{1.3}}$  &      ${{92.3}}_{\textcolor{blue}{3.3}}$ & ${{89.3}}_{\textcolor{blue}{2.4}}$
& ${{79.1}}_{\textcolor{blue}{1.5}}$
& ${{95.6}}_{\textcolor{blue}{2.3}}$ & ${{88.4}}_{\textcolor{blue}{1.9}}$ & ${{64.6}}_{\textcolor{blue}{2.8}}$ & ${82.9}$   \\ \cline{2-13}

& Vanilla ICL & \multirow{2}{*}{8}             &  ${96.8}_{\textcolor{blue}{7.3}}$     &   ${34.3}_{\textcolor{blue}{5.4}}$    &  ${93.4}_{\textcolor{blue}{5.8}}$  &  ${92.7}_{\textcolor{blue}{6.4}}$  & ${83.5}_{\textcolor{blue}{5.5}}$     &   ${66.9}_{\textcolor{blue}{4.8}}$   & ${84.1}_{\textcolor{blue}{6.2}}$        &    ${84.7}_{\textcolor{blue}{5.5}}$    & ${62.0}_{\textcolor{blue}{5.2}}$  & 77.6 \\ 
& IDAICL &         &  $96.9_{\textcolor{blue}{2.3}}$     & $50.3_{\textcolor{blue}{1.5}}$      & $93.9_{\textcolor{blue}{2.2}}$   & ${93.0}_{\textcolor{blue}{1.4}}$   &   $89.0_{\textcolor{blue}{1.0}}$      &    ${83.1}_{\textcolor{blue}{1.7}}$  &  $95.9_{\textcolor{blue}{2.0}}$       &     $88.0_{\textcolor{blue}{1.2}}$   & $70.4_{\textcolor{blue}{1.8}}$ & 84.5 \\ 
\bottomrule[1.5pt]
\end{tabular}}
\caption{Comparison results of Macro-F1 between Vanilla ICL and IDAICL under varying values of $m$ on the LLaMA models with 13B and 33B parameters.}
\label{llamajiben}
\end{table*}

\begin{table*}[t]
% \small
\centering
\resizebox{1\linewidth}{!}{
\begin{tabular}{l|l|c|cccccccccc}
\toprule[1.5pt]
\rowcolor{gray!30}
             PLM & Method & m  & SST-2 & SST-5 & MR & CR & Amazon & Subj & TREC & DBPedia & AGNews & CB \\ \midrule\midrule
\multirow{8}{*}{\rotatebox{90}{GPT-2 0.1B}} 
 % & Vanilla ICL & \multirow{2}{*}{1}             &  ${45.6}_{\textcolor{blue}{5.8}}$     &   ${20.9}_{\textcolor{blue}{6.7}}$    &  ${40.2}_{\textcolor{blue}{6.1}}$  &  ${47.3}_{\textcolor{blue}{5.9}}$  & $27.0_{\textcolor{blue}{7.1}}$     & ${37.8}_{\textcolor{blue}{5.2}}$     &   ${26.7}_{\textcolor{blue}{4.4}}$   & ${34.7}_{\textcolor{blue}{5.9}}$        &    ${28.6}_{\textcolor{blue}{9.1}}$    & ${35.2}_{\textcolor{blue}{4.3}}$   \\ 
% & IDAICL &         &  $56.2_{\textcolor{blue}{3.1}}$     & $28.5_{\textcolor{blue}{2.7}}$      & $58.1_{\textcolor{blue}{1.8}}$   & ${58.0}_{\textcolor{blue}{2.5}}$   &   $38.6_{\textcolor{blue}{3.4}}$   & $50.3_{\textcolor{blue}{1.9}}$      &    ${41.6}_{\textcolor{blue}{2.2}}$  &  $66.5_{\textcolor{blue}{1.8}}$       &     $51.2_{\textcolor{blue}{1.9}}$   & $59.0_{\textcolor{blue}{2.1}}$   \\ \cline{2-13}
& Vanilla ICL & \multirow{2}{*}{4}             &  ${56.3}_{\textcolor{blue}{7.1}}$     &   ${28.4}_{\textcolor{blue}{8.8}}$    &  ${55.4}_{\textcolor{blue}{7.4}}$  &  ${54.2}_{\textcolor{blue}{6.2}}$  & $30.8_{\textcolor{blue}{8.4}}$     & ${52.9}_{\textcolor{blue}{7.9}}$     &   ${32.2}_{\textcolor{blue}{5.1}}$   & ${44.3}_{\textcolor{blue}{6.2}}$        &    ${42.8}_{\textcolor{blue}{9.3}}$    & ${42.1}_{\textcolor{blue}{9.6}}$   \\ 
& IDAICL &         &  $69.5_{\textcolor{blue}{2.6}}$     & $35.3_{\textcolor{blue}{1.1}}$      & $66.4_{\textcolor{blue}{2.3}}$   & ${67.2}_{\textcolor{blue}{2.7}}$   &   $39.3_{\textcolor{blue}{2.9}}$   & $57.2_{\textcolor{blue}{2.6}}$      &    ${44.3}_{\textcolor{blue}{1.8}}$  &  $62.2_{\textcolor{blue}{2.3}}$       &     $65.5_{\textcolor{blue}{2.7}}$   & $49.2_{\textcolor{blue}{1.9}}$   \\ \cline{2-13}

& Vanilla ICL & \multirow{2}{*}{8}             &  ${60.8}_{\textcolor{blue}{8.3}}$     &   ${30.6}_{\textcolor{blue}{6.9}}$    &  ${57.5}_{\textcolor{blue}{9.7}}$  &  ${56.0}_{\textcolor{blue}{5.1}}$  & $33.6_{\textcolor{blue}{7.8}}$     & ${53.7}_{\textcolor{blue}{5.6}}$     &   ${33.0}_{\textcolor{blue}{10.7}}$   & ${52.1}_{\textcolor{blue}{5.8}}$        &    ${45.6}_{\textcolor{blue}{9.1}}$    & ${45.4}_{\textcolor{blue}{6.2}}$   \\ 
& IDAICL &         &  $71.4_{\textcolor{blue}{1.8}}$     & $36.1_{\textcolor{blue}{2.9}}$      & $67.6_{\textcolor{blue}{1.8}}$   & ${68.6}_{\textcolor{blue}{2.2}}$   &   $40.0_{\textcolor{blue}{0.7}}$   & $58.5_{\textcolor{blue}{2.5}}$      &    ${45.6}_{\textcolor{blue}{1.9}}$  &  $63.6_{\textcolor{blue}{1.1}}$       &     $66.9_{\textcolor{blue}{1.6}}$   & $50.6_{\textcolor{blue}{2.7}}$   \\ \cline{2-13}

& Vanilla ICL   & \multirow{2}{*}{12}             &  ${64.5}_{\textcolor{blue}{6.0}}$     &  ${30.8}_{\textcolor{blue}{7.1}}$     & ${59.3}_{\textcolor{blue}{5.6}}$   & ${59.1}_{\textcolor{blue}{8.4}}$   & $33.9_{\textcolor{blue}{5.5}}$     & ${56.6}_{\textcolor{blue}{8.9}}$     & ${35.8}_{\textcolor{blue}{7.1}}$     & ${52.3}_{\textcolor{blue}{11.4}}$        & ${47.4}_{\textcolor{blue}{6.0}}$       & ${47.4}_{\textcolor{blue}{7.7}}$    \\ 
& IDAICL     &     &  $72.2_{\textcolor{blue}{1.1}}$     & $36.7_{\textcolor{blue}{2.2}}$      & $70.1_{\textcolor{blue}{1.7}}$   & $69.3_{\textcolor{blue}{1.8}}$   & $40.8_{\textcolor{blue}{1.2}}$     & $60.9_{\textcolor{blue}{1.5}}$     & $47.0_{\textcolor{blue}{2.7}}$     & $65.5_{\textcolor{blue}{1.9}}$        & $67.8_{\textcolor{blue}{2.2}}$       & $51.2_{\textcolor{blue}{3.3}}$   \\ \cline{2-13}
& Vanilla ICL   & \multirow{2}{*}{16}             &  ${64.3}_{\textcolor{blue}{6.1}}$     &  ${33.5}_{\textcolor{blue}{7.1}}$     & ${59.9}_{\textcolor{blue}{6.6}}$   & ${61.7}_{\textcolor{blue}{7.5}}$   & $34.6_{\textcolor{blue}{6.9}}$     & ${56.1}_{\textcolor{blue}{6.2}}$     & ${36.9}_{\textcolor{blue}{5.7}}$     & ${54.1}_{\textcolor{blue}{7.2}}$        & ${47.9}_{\textcolor{blue}{8.0}}$       & ${48.9}_{\textcolor{blue}{7.7}}$    \\ 
& IDAICL     &     &  $72.9_{\textcolor{blue}{2.5}}$     & $38.0_{\textcolor{blue}{2.4}}$      & $69.7_{\textcolor{blue}{1.3}}$   & $69.9_{\textcolor{blue}{2.1}}$   & $41.7_{\textcolor{blue}{0.9}}$     & $60.6_{\textcolor{blue}{1.1}}$     & $46.6_{\textcolor{blue}{1.9}}$     & $65.9_{\textcolor{blue}{2.6}}$        & $65.7_{\textcolor{blue}{1.0}}$       & $51.8_{\textcolor{blue}{2.2}}$   \\ 
\midrule\midrule
\multirow{8}{*}{\rotatebox{90}{GPT-2 0.3B}} 
% & Vanilla ICL & \multirow{2}{*}{1}             &  ${44.6}_{\textcolor{blue}{6.9}}$     &   ${19.8}_{\textcolor{blue}{8.2}}$    &  ${40.1}_{\textcolor{blue}{5.6}}$  &  ${46.0}_{\textcolor{blue}{4.7}}$  & $29.8_{\textcolor{blue}{5.1}}$     & ${48.9}_{\textcolor{blue}{5.3}}$     &   ${25.6}_{\textcolor{blue}{7.1}}$   & ${35.8}_{\textcolor{blue}{8.2}}$        &    ${26.9}_{\textcolor{blue}{4.9}}$    & ${30.6}_{\textcolor{blue}{8.1}}$   \\ 
% & IDAICL &         &  $70.2_{\textcolor{blue}{1.9}}$     & $28.8_{\textcolor{blue}{2.3}}$      & $62.1_{\textcolor{blue}{2.4}}$   & ${59.1}_{\textcolor{blue}{1.7}}$   &   $39.6_{\textcolor{blue}{2.5}}$   & $61.1_{\textcolor{blue}{3.2}}$      &    ${48.7}_{\textcolor{blue}{2.9}}$  &  $67.0_{\textcolor{blue}{3.3}}$       &     $48.1_{\textcolor{blue}{3.5}}$   & $56.9_{\textcolor{blue}{1.9}}$   \\ \cline{2-13}
& Vanilla ICL & \multirow{2}{*}{4}             &  ${60.8}_{\textcolor{blue}{7.5}}$     &   ${26.6}_{\textcolor{blue}{6.8}}$    &  ${50.5}_{\textcolor{blue}{7.1}}$  &  ${52.3}_{\textcolor{blue}{6.1}}$  & $30.5_{\textcolor{blue}{5.2}}$     & ${53.2}_{\textcolor{blue}{8.3}}$     &   ${32.8}_{\textcolor{blue}{8.1}}$   & ${50.5}_{\textcolor{blue}{4.8}}$        &    ${41.3}_{\textcolor{blue}{5.9}}$    & ${42.7}_{\textcolor{blue}{7.1}}$   \\ 
& IDAICL &         &  $78.4_{\textcolor{blue}{1.7}}$     & $33.1_{\textcolor{blue}{2.5}}$      & $66.6_{\textcolor{blue}{0.9}}$   & ${70.3}_{\textcolor{blue}{2.3}}$   &   $40.1_{\textcolor{blue}{1.5}}$   & $69.4_{\textcolor{blue}{1.7}}$      &    ${45.6}_{\textcolor{blue}{3.3}}$  &  $66.2_{\textcolor{blue}{2.1}}$       &     $62.8_{\textcolor{blue}{3.7}}$   & $50.4_{\textcolor{blue}{1.8}}$   \\ \cline{2-13}

& Vanilla ICL & \multirow{2}{*}{8}             &  ${58.9}_{\textcolor{blue}{8.7}}$     &   ${29.4}_{\textcolor{blue}{6.1}}$    &  ${52.4}_{\textcolor{blue}{8.9}}$  &  ${54.8}_{\textcolor{blue}{8.2}}$  & $32.7_{\textcolor{blue}{7.9}}$     & ${53.5}_{\textcolor{blue}{6.7}}$     &   ${34.0}_{\textcolor{blue}{8.2}}$   & ${59.1}_{\textcolor{blue}{9.7}}$        &    ${43.8}_{\textcolor{blue}{6.4}}$    & ${46.9}_{\textcolor{blue}{7.6}}$   \\ 
& IDAICL &         &  $80.8_{\textcolor{blue}{1.7}}$     & $34.8_{\textcolor{blue}{1.9}}$      & $69.5_{\textcolor{blue}{1.1}}$   & ${71.5}_{\textcolor{blue}{0.8}}$   &   $41.5_{\textcolor{blue}{1.7}}$   & $70.3_{\textcolor{blue}{2.6}}$      &    ${46.2}_{\textcolor{blue}{2.2}}$  &  $68.1_{\textcolor{blue}{1.7}}$       &     $63.3_{\textcolor{blue}{2.1}}$   & $51.5_{\textcolor{blue}{2.5}}$   \\ \cline{2-13}
        
        & Vanilla ICL & \multirow{2}{*}{12}           &  ${62.9}_{\textcolor{blue}{14.4}}$     &   ${30.6}_{\textcolor{blue}{7.8}}$    &  ${55.2}_{\textcolor{blue}{6.2}}$  &  ${56.1}_{\textcolor{blue}{6.7}}$  & $34.2_{\textcolor{blue}{7.5}}$     & ${56.8}_{\textcolor{blue}{7.1}}$     &   ${36.2}_{\textcolor{blue}{9.8}}$   & ${58.0}_{\textcolor{blue}{7.3}}$        &    ${46.5}_{\textcolor{blue}{9.3}}$    & ${48.6}_{\textcolor{blue}{6.6}}$   \\ 
& IDAICL &         &  $82.2_{\textcolor{blue}{2.3}}$     & $36.1_{\textcolor{blue}{1.8}}$      & $68.9_{\textcolor{blue}{2.4}}$   & ${72.0}_{\textcolor{blue}{1.5}}$   &   $43.7_{\textcolor{blue}{0.6}}$   & $71.4_{\textcolor{blue}{2.4}}$      &    ${48.3}_{\textcolor{blue}{1.3}}$  &  $70.5_{\textcolor{blue}{1.9}}$       &     $65.2_{\textcolor{blue}{2.2}}$   & $52.9_{\textcolor{blue}{1.4}}$   \\ \cline{2-13}
& Vanilla ICL   & \multirow{2}{*}{16}             &  ${67.4}_{\textcolor{blue}{6.3}}$     &  ${31.7}_{\textcolor{blue}{7.1}}$     & ${57.6}_{\textcolor{blue}{8.6}}$   & ${56.6}_{\textcolor{blue}{5.2}}$   & $34.7_{\textcolor{blue}{6.2}}$     & ${57.0}_{\textcolor{blue}{5.3}}$     & ${38.1}_{\textcolor{blue}{6.9}}$     & ${59.3}_{\textcolor{blue}{8.2}}$        & ${45.2}_{\textcolor{blue}{7.6}}$       & ${49.4}_{\textcolor{blue}{8.7}}$    \\ 
& IDAICL     &     &  $81.5_{\textcolor{blue}{2.8}}$     & $36.8_{\textcolor{blue}{1.2}}$      & $70.4_{\textcolor{blue}{1.7}}$   & $72.9_{\textcolor{blue}{2.1}}$   & $43.1_{\textcolor{blue}{1.3}}$     & $71.9_{\textcolor{blue}{2.7}}$     & $48.7_{\textcolor{blue}{1.1}}$     & $70.9_{\textcolor{blue}{2.9}}$        & $65.8_{\textcolor{blue}{1.2}}$       & $52.4_{\textcolor{blue}{1.8}}$   \\ 
\bottomrule[1.5pt]
\end{tabular}}
\caption{Accuracy comparison between Vanilla ICL and IDAICL under varying values of $m$ on the GPT-2 models with 0.1B and 0.3B parameters.}
\label{gptsmall}
\end{table*}

% \begin{figure*}[t] 
% \centering
% % \vspace{-0.1in}
% % \includesvg[width=0.99\textwidth]{fig71117.svg}
% \includesvg[width=0.99\textwidth]{pic/fig71216.svg}
% % \vspace{-0.12in}
% \caption{Comparison results among Vanilla ICL, IDAICL (pre-inference), and IDAICL (online) on eight datasets. The GPT-2 model with 0.8B parameters is employed for these comparisons. 
% The performance of IDAICL (online) surpasses Vanilla ICL significantly, yet it falls short compared to IDAICL (pre-inference).
% }
% \label{online}
% % \vspace{-0.02in}
% % \vspace{-0.16in}
% \end{figure*}

\begin{figure*}[t] 
\centering
% \vspace{-0.05in}
% \includegraphics[width=0.48\textwidth]{sense927.pdf}
\includegraphics[width=1\textwidth]{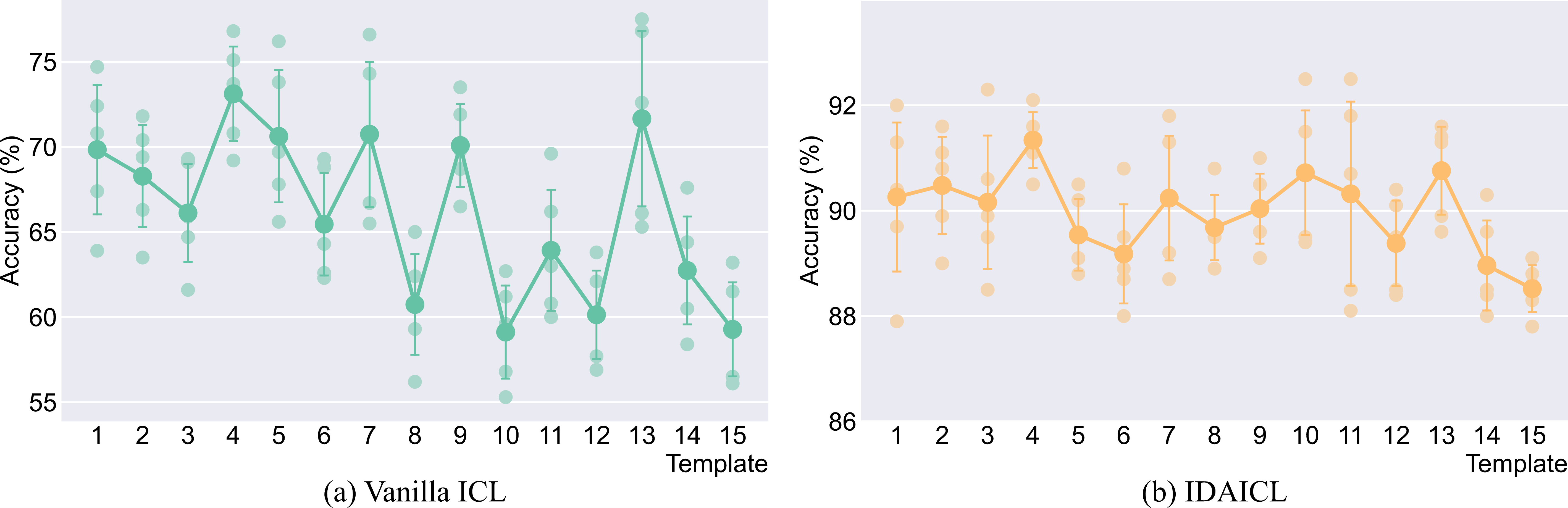}
% \vspace{-0.05in}
% \caption{(a) and (b): Analysis of $\lambda$ on SST-2 and SUBJ. (c): Analysis of $\tau$ on SST-5 and Amazon.}
\caption{Comparison results between Vanilla ICL and IDAICL across fifteen templates. The evaluation is conducted using the GPT-2 model with 1.5B parameters. The performance of IDAICL exceeds that of Vanilla ICL and demonstrates greater robustness across various templates.}
\label{tempfulu}
% \vspace{-0.02in}
% \vspace{-0.16in}
\end{figure*}

\begin{table*}[t]
\small
% \vspace{-0.02in}
\centering
% \resizebox{1\linewidth}{!}{
\begin{tabular}{l|l|cccccccc}
\toprule[1.2pt]
\rowcolor{gray!30}
      PLM &   Method       & SST-5 & MR & AGNews &  TREC & SST-2 & Subj & DBPedia & Avg. \\ \midrule\midrule
\multirow{5}{*}{{GPT-2 1.5B}}  & Vanilla ICL             & ${30.8}_{\textcolor{blue}{6.1}}$ & ${64.9}_{\textcolor{blue}{8.3}}$ & ${57.5}_{\textcolor{blue}{6.7}}$  & ${40.4}_{\textcolor{blue}{5.1}}$ & ${57.2}_{\textcolor{blue}{7.0}}$ & ${57.3}_{\textcolor{blue}{10.3}}$ & ${67.6}_{\textcolor{blue}{7.5}}$ & 53.7\\
& ConCa     &   ${32.8}_{\textcolor{blue}{7.1}}$    &   $74.5_{\textcolor{blue}{5.1}}$ &  $62.7_{\textcolor{blue}{6.1}}$          &  ${45.8}_{\textcolor{blue}{2.5}}$  & ${73.9}_{\textcolor{blue}{8.6}}$ & ${68.3}_{\textcolor{blue}{7.4}}$ & ${75.0}_{\textcolor{blue}{4.0}}$ & 61.9 \\
& ${\text{P}}$$\scriptsize{\text{RO}}$${\text{C}}$$\scriptsize{\text{A}}^{*}$       &    ${\underline{36.5}}_{\textcolor{blue}{4.4}}$    &   ${{80.8}}_{\textcolor{blue}{6.4}}$ & ${75.5}_{\textcolor{blue}{3.2}}$      &  ${46.0}_{\textcolor{blue}{2.5}}$ & ${\underline{88.0}}_{\textcolor{blue}{1.3}}$ & ${\boldsymbol{80.2}}_{\textcolor{blue}{3.3}}$ & ${\underline{89.4}}_{\textcolor{blue}{0.7}}$ & $\underline{70.9}$ \\ 
% ${\text{PMI}}_{\text{DC}}$         & 
% $\underline{37.3}_{\textcolor{blue}{4.5}}$     & 
% ${76.8}_{\textcolor{blue}{2.5}}$   &   ${70.7}_{\textcolor{blue}{4.2}}$       &  
% ${46.9}_{\textcolor{blue}{3.7}}$ &   
% ${74.8}_{\textcolor{blue}{2.5}}$ &  $66.5_{\textcolor{blue}{5.9}}$&  $73.2_{\textcolor{blue}{8.9}}$ & 63.7 \\
& D-ConCa     & ${{31.7}}_{\textcolor{blue}{3.3}}$       &    ${\underline{80.9}}_{\textcolor{blue}{3.7}}$   &   ${\underline{77.0}}_{\textcolor{blue}{4.1}}$ &  ${\underline{47.1}}_{\textcolor{blue}{2.8}}$      & ${86.5}_{\textcolor{blue}{4.4}}$  & ${76.8}_{\textcolor{blue}{5.2}}$ & ${86.1}_{\textcolor{blue}{6.3}}$ & 69.4\\

% SimCSE     &       &       &    &    &      &      &      &         &        &    \\
% IDA+SimCSE &       &       &    &    &      &      &      &         &        &     \\ \hline
& IDAICL        & ${\boldsymbol{40.8}}_{\textcolor{blue}{1.9}}$      &   ${\boldsymbol{82.1}}_{\textcolor{blue}{1.2}}$ &  ${\boldsymbol{80.8}}_{\textcolor{blue}{2.4}}$  &      ${\boldsymbol{52.0}}_{\textcolor{blue}{2.5}}$ & ${\boldsymbol{89.5}}_{\textcolor{blue}{1.8}}$
& ${\underline{80.1}}_{\textcolor{blue}{2.9}}$
& ${\boldsymbol{91.0}}_{\textcolor{blue}{2.5}}$ & $\boldsymbol{73.8}$\\ \midrule\midrule
\multirow{5}{*}{{GPT-Neo}}  & Vanilla ICL             & ${31.5}_{\textcolor{blue}{6.4}}$ & ${70.6}_{\textcolor{blue}{8.1}}$ & ${71.9}_{\textcolor{blue}{6.8}}$  & ${53.0}_{\textcolor{blue}{6.9}}$ & ${74.9}_{\textcolor{blue}{8.3}}$ & ${57.9}_{\textcolor{blue}{6.3}}$ & ${78.5}_{\textcolor{blue}{6.5}}$ & 62.6 \\
& ConCa     &   ${33.9}_{\textcolor{blue}{4.3}}$    &   $78.2_{\textcolor{blue}{5.3}}$ &  $73.6_{\textcolor{blue}{3.8}}$          &  ${55.9}_{\textcolor{blue}{7.2}}$  & ${82.0}_{\textcolor{blue}{9.5}}$ & ${71.3}_{\textcolor{blue}{6.4}}$ & ${90.0}_{\textcolor{blue}{3.6}}$ & 69.3 \\
& ${\text{P}}$$\scriptsize{\text{RO}}$${\text{C}}$$\scriptsize{\text{A}}^{*}$       &    ${\underline{39.4}}_{\textcolor{blue}{4.0}}$    &   ${{77.8}}_{\textcolor{blue}{13.9}}$ & ${78.9}_{\textcolor{blue}{2.5}}$      &  ${56.0}_{\textcolor{blue}{3.6}}$ & ${\boldsymbol{91.9}}_{\textcolor{blue}{1.2}}$ & ${\underline{81.3}}_{\textcolor{blue}{3.8}}$ & ${\underline{92.0}}_{\textcolor{blue}{1.5}}$ & $\underline{73.9}$\\ 
% ${\text{PMI}}_{\text{DC}}$         & 
% $\underline{37.3}_{\textcolor{blue}{4.5}}$     & 
% ${76.8}_{\textcolor{blue}{2.5}}$   &   ${70.7}_{\textcolor{blue}{4.2}}$       &  
% ${46.9}_{\textcolor{blue}{3.7}}$ &   
% ${74.8}_{\textcolor{blue}{2.5}}$ &  $66.5_{\textcolor{blue}{5.9}}$&  $73.2_{\textcolor{blue}{8.9}}$ & 63.7 \\
& D-ConCa     & ${{32.9}}_{\textcolor{blue}{4.1}}$       &    ${\underline{84.6}}_{\textcolor{blue}{2.8}}$   &   ${\underline{81.2}}_{\textcolor{blue}{3.9}}$ &  ${\underline{57.6}}_{\textcolor{blue}{4.7}}$      & ${\underline{91.6}}_{\textcolor{blue}{5.3}}$  & ${70.9}_{\textcolor{blue}{2.9}}$ & ${85.7}_{\textcolor{blue}{3.1}}$ & 72.1\\

% SimCSE     &       &       &    &    &      &      &      &         &        &    \\
% IDA+SimCSE &       &       &    &    &      &      &      &         &        &     \\ \hline
& IDAICL        & ${\boldsymbol{42.2}}_{\textcolor{blue}{2.5}}$      &   ${\boldsymbol{85.9}}_{\textcolor{blue}{1.6}}$ &  ${\boldsymbol{83.1}}_{\textcolor{blue}{1.9}}$  &      ${\boldsymbol{61.4}}_{\textcolor{blue}{1.7}}$ & ${{91.2}}_{\textcolor{blue}{2.4}}$
& ${\boldsymbol{82.3}}_{\textcolor{blue}{3.1}}$
& ${\boldsymbol{93.0}}_{\textcolor{blue}{1.5}}$ &  $\boldsymbol{77.0}$\\ 
% Optimal        & $\boldsymbol{43.8}_{\textcolor{blue}{2.9}}$      &   $\boldsymbol{85.4}_{\textcolor{blue}{3.1}}$ &  $\boldsymbol{84.6}_{\textcolor{blue}{2.1}}$  &      $\boldsymbol{63.7}_{\textcolor{blue}{4.2}}$ & $\boldsymbol{90.5}_{\textcolor{blue}{3.4}}$ & $\boldsymbol{81.2}_{\textcolor{blue}{2.7}}$ & $\boldsymbol{91.3}_{\textcolor{blue}{1.9}}$   \\
\bottomrule[1.2pt]
\end{tabular}
% }
% \vspace{-0.05in}
\caption{Accuracy comparison between IDAICL and other prediction calibration approaches using the GPT-2 (with 1.5B parameters) and GPT-Neo models, with $m$ setting to 8. The templates used align with those utilized by Han et al.~\shortcite{han2022prototypical}. 
$*$ indicates that the results were derived from the original paper.
% ``Optimal" refers to the optimal performance achievable when combining the IDAICL method with other approaches. 
}
\label{tabcalibrating}
% \vspace{-0.1in}
\end{table*}

\begin{figure}[t] 
\centering
% \vspace{-0.03in}
% \includegraphics[width=0.48\textwidth]{sense927.pdf}
\includegraphics[width=0.48\textwidth]{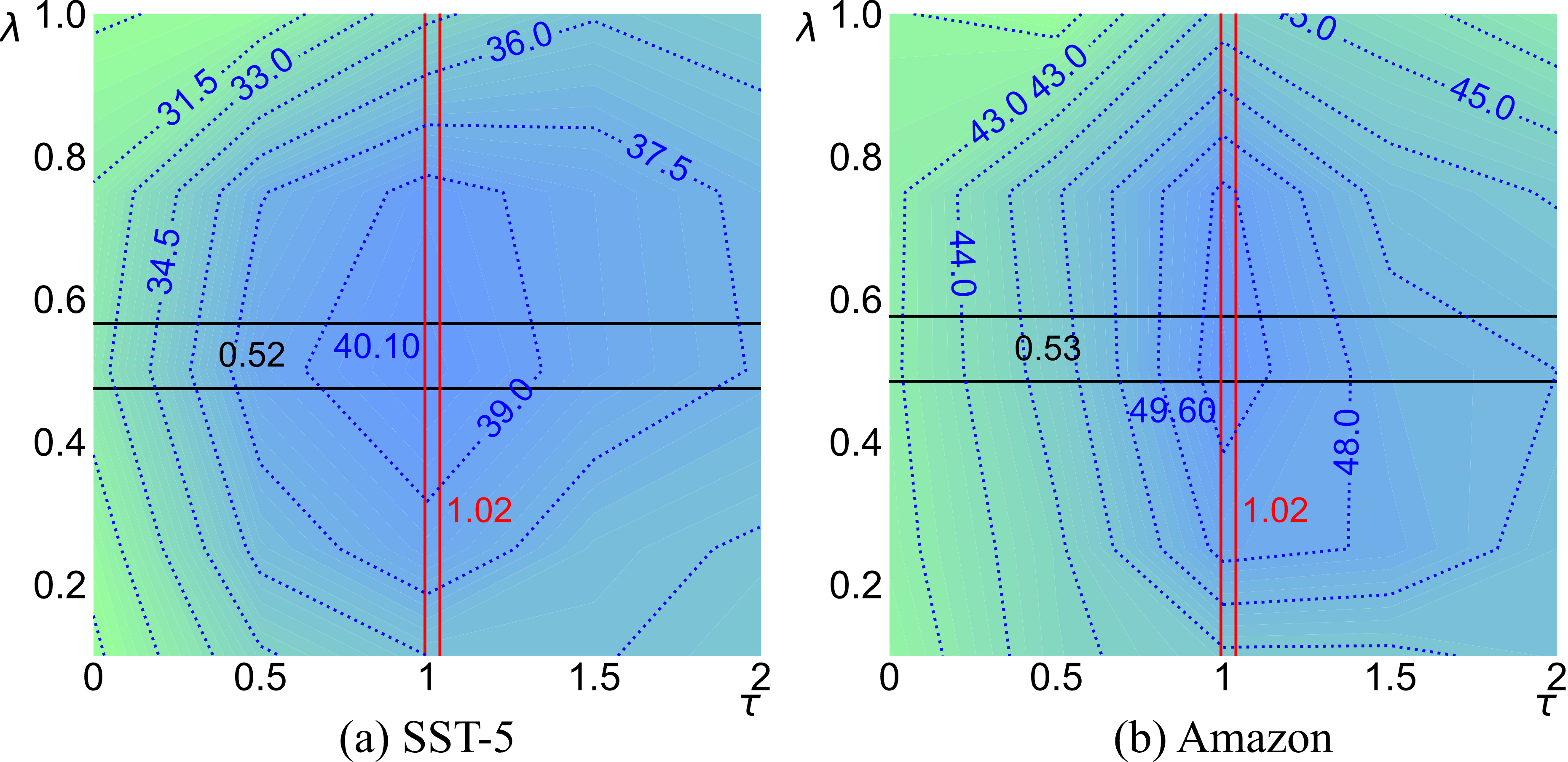}
% \vspace{-0.05in}
% \caption{(a) and (b): Analysis of $\lambda$ on SST-2 and SUBJ. (c): Analysis of $\tau$ on SST-5 and Amazon.}
\caption{Results of sensitivity tests for two hyperparameters within IDAICL, i.e., $\lambda$ and $\tau$, using the GPT-2 model with 0.8B parameters, with $m$ setting to 12. Optimal performance is achieved when $\lambda\approx0.5$ and $\tau\approx1$. 
}
\label{sense}
% \vspace{-0.02in}
% \vspace{-0.16in}
\end{figure}

\section{Details of Compared Baselines}
\label{method}
The compared methods are described as follows:
\begin{itemize}
    \item \textbf{Vanilla ICL}: We use the PLMs as they are and implement ICL by conditioning it on a concatenation of $m$ training examples, following the approach outlined by Brown et al.~\shortcite{brown2020language}.
\item \textbf{MetaICL}: The fundamental concept underlying MetaICL is to utilize a multi-task learning framework across a diverse range of meta-training tasks~\cite{min-etal-2022-metaicl}.
\item \textbf{Channel ICL}: It employs a noisy channel approach for language model prompting in few-shot text classification~\cite{min-etal-2022-noisy}.

\item \textbf{EPR}: It employs language models to autonomously label examples that are suitable as effective prompts and subsequently trains a prompt retriever based on this acquired signal~\cite{rubin-etal-2022-learning}.

\item \textbf{ConCa}: It assesses the model's inclination towards specific answers by introducing a dummy test input that lacks content~\cite{zhao2021calibrate}.

    \item \textbf{$\text{P}$$\small{\text{RO}}$$\text{C}$$\small{\text{A}}$}: The prediction of $\text{P}$$\small{\text{RO}}$$\text{C}$$\small{\text{A}}$ is calibrated based on the likelihood of prototypical clusters~\cite{han2022prototypical}.

\item \textbf{D-ConCa}: It initially assesses the impacts of various label biases by employing randomly sampled words from the task corpus. During inference, it utilizes the estimated label bias to calibrate the model's output probabilities~\cite{fei-etal-2023-mitigating}.

\end{itemize}

\section{More Details of Experimental Settings}
\label{setting}

The entire implementation is conducted utilizing PyTorch~\cite{paszke2019pytorch} and Transformers~\cite{wolf-etal-2020-transformers}. We follow the parameter configurations and details specified in previous research~\cite{min-etal-2022-noisy}. The number of demonstrations is primarily set to $m=12$, but we also explore $m$ values of $\{1, 4, 8, 12,16\}$ in the ablations, with the specific settings detailed in the respective sections. Demonstration examples for each test sample are randomly selected from the training data, unless specific methods employ a specially designed selection method, such as EPR~\cite{rubin-etal-2022-learning}. The values of the feature mean and covariance matrix are estimated from the demonstration set containing demonstration examples corresponding to all test samples. We depart from the assumption made in previous studies, which presupposes an equal distribution of training examples across all classes~\cite{gao-etal-2021-making,logan-iv-etal-2022-cutting}, in order to facilitate a more realistic and demanding evaluation.

% was selected from the set \{0.1, 0.25, 0.5, 0.75, 1\}, and the value of
Each experiment is repeated under five different random seeds. 
The batch size is set to 32, and the sequence length is configured to 128 for datasets with shorter texts, including SST-2~\cite{socher-etal-2013-recursive}, SST-5~\cite{socher-etal-2013-recursive}, MR~\cite{pang-lee-2005-seeing}, CR~\cite{hu2004mining}, and TREC~\cite{voorhees2000building}. On the other hand, for datasets with longer input texts, including AGNews~\cite{zhang2015character}, DBPedia~\cite{lehmann2015dbpedia}, Subj~\cite{pang-lee-2004-sentimental}, CB~\cite{de2019commitmentbank}, and Amazon~\cite{mcauley2013hidden}, a batch size of 16 and a sequence length of 256 are employed. Regarding the hyperparameters in IDAICL, the values of $\lambda$  
and $\tau$ are fixed at 0.5 and 1, respectively, except in sensitivity tests. The settings used for the compared methods adhere to those specified in the original papers~\cite{min-etal-2022-noisy,min-etal-2022-metaicl,rubin-etal-2022-learning,zhao2021calibrate,han2022prototypical,fei-etal-2023-mitigating}. 
% For EPR, we trained different dense retrievers using distinct random seeds. 
Accuracy serves as the primary evaluation metric, alongside the provided values of Macro-F1 for the LLaMA model. For each task, a specific template is utilized for inference, as detailed in Table~\ref{temp}. Additionally, we also examine the impact of different templates on the performance of IDAICL following those outlined by Zhao et al.~\shortcite{zhao2021calibrate}, which include question-answer templates, conversation-style templates, prompts resembling web pages, and variations on label names, as listed in Table~\ref{formulate}.

% With the exception of the templates outlined in the original paper of GlobalE and LocalE~\cite{lu-etal-2022-fantastically}, we opt for a minimal format to convert the input into a sequence. This entails concatenating multiple inputs and utilizing the label words directly from each dataset, aligning with the conventions established in previous studies~\cite{ye-etal-2021-crossfit,min-etal-2022-metaicl,logan-iv-etal-2022-cutting}.

\section{More Comparison Results}
\label{comp}
% \label{sec:appendix}
The comparison results between Vanilla ICL and IDAICL on LLaMA models with 13B and 33B parameters across various datasets are presented in Table~\ref{llamajiben}. Additionally, the corresponding results for GPT-2 models with 0.1B and 0.3B parameters are outlined in Table~\ref{gptsmall}. It is evident that IDAICL consistently outperforms Vanilla ICL across all datasets and different model sizes, highlighting the high generalization capability of IDAICL. Additionally, IDAICL showcases reduced performance variance and significantly enhances the worst-case performance. 
Based on the findings presented in Table~\ref{tabcalibrating}, IDAICL generally outperforms other prediction calibration methods, demonstrating the significance of statistical properties derived from the input data distribution in the predictions of PLMs. 

\begin{figure*}[t] 
\centering
% \vspace{-0.03in}
\includegraphics[width=1\textwidth]{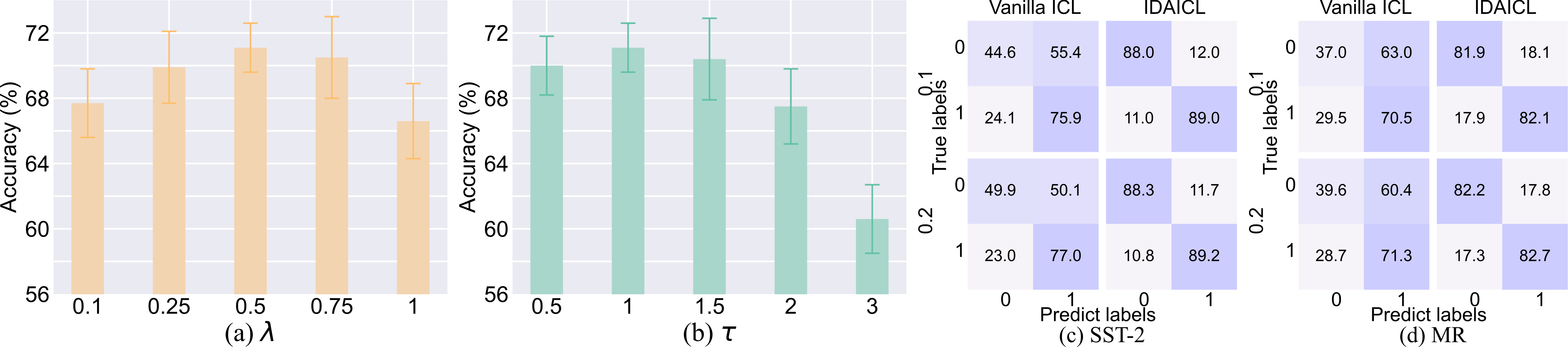}
% \vspace{-0.12in}
\caption{
(a) and (b): Average accuracy across ten datasets for various values of $\lambda$ and $\tau$. Optimal average performance is attained when $\lambda=0.5$ and $\tau=1$. (c) and (d): Confusion matrices for the SST-2 and MR datasets under two levels of imbalance, where the proportions of the negative class in demonstrations are set to 0.1 and 0.2, respectively. When compared to Vanilla ICL, IDAICL improves the performance of the minor class. These experiments are conducted on the GPT-2 model with 1.5B parameters, setting $m$ to 12. 
}
\label{imbfulu1026}
% \vspace{-0.02in}
% \vspace{-0.16in}
\end{figure*}

\begin{figure*}[t] 
\centering
% \vspace{-0.05in}
% \includegraphics[width=0.48\textwidth]{sense927.pdf}
\includegraphics[width=1\textwidth]{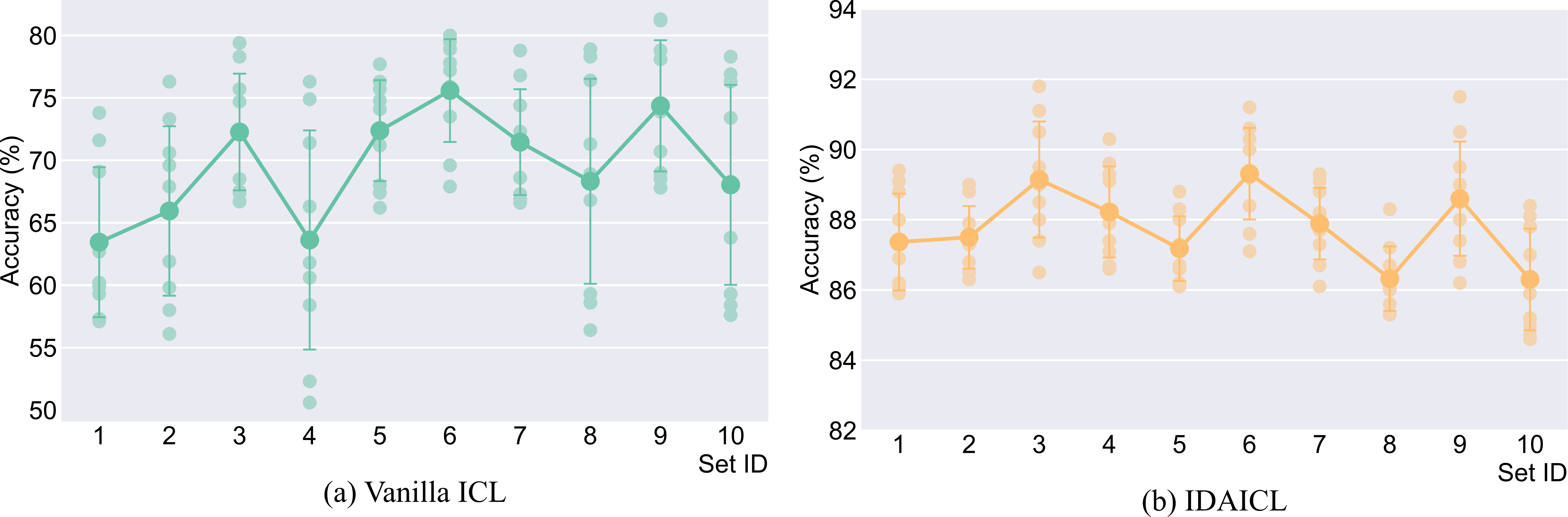}
\caption{Comparison results between Vanilla ICL and IDAICL across various demonstrations and permutations. The GPT-2 model with 0.8B parameters is employed for this analysis, setting $m$ to 12. 
IDAICL exhibits smaller performance variance across different demonstrations and permutations compared to Vanilla ICL.
% compared with 
 }
\label{shunxu}
% \vspace{-0.02in}
% \vspace{-0.16in}
\end{figure*}

\section{More Results for Varying Templates}
The comparison results between Vanilla ICL and IDAICL under all fifteen prompt templates are presented in Figure~\ref{tempfulu}, illustrating that IDAICL consistently enhances both average and worst-case accuracy across all templates. Furthermore, the performance variance of IDAICL among different templates is notably smaller when compared to Vanilla ICL, highlighting the robustness of IDAICL's performance across diverse templates.

\section{More Sensitivity and Ablation Studies}
\label{abafulusection}

We performed sensitivity tests on two hyperparameters within IDAICL: $\lambda$ and $\tau$. These values govern the strength of implicit augmentation and the influence of the class proportion term, respectively.
As depicted in Figure~\ref{sense}, optimal performance is achieved when $\lambda\!\approx\!0.5$ and $\tau\!\approx\!1$ for both datasets. 
Furthermore, Figures~\ref{imbfulu1026}(a) and (b) illustrate the average performance of ten datasets across different hyperparameter settings. 
% These experiments are conducted on GPT-2 with 1.5B parameters setting $m$ to 12. 
Much like the earlier findings, the best average performance is identified at $\lambda\!=\!0.5$ and $\tau\!=\!1$. Consequently, setting $\lambda$ as 0.5 and $\tau$ as 1 is recommended for real applications. 
Furthermore, the performance remains stable within the ranges of $\lambda \in \{0.25, 0.5, 0.75\}$ and $\tau \in \{0.5,1,1.5\}$, indicating that adjustments can be made within these stable ranges.

\section{More Results for Imbalanced Labels}

% We have also assessed the performance of IDAICL across different numbers of demonstrations using the GPT-2 model with 1.5B parameters. As illustrated by the findings in Table~\ref{shot}, IDAICL consistently demonstrates superior performance across varying numbers of demonstrations. This performance improvement is noteworthy even when the value of $m$ is small. This is attributed to the fact that, even in scenarios with small $m$ values, our augmentation strategy is capable of conveying a large amount of information to PLMs.
% We evaluated the performance of IDAICL on two five-class classification datasets, including SST-5 and Amazon, with class proportions in the demonstrations setting to 0.05, 0.05, 0.1, 0.3, and 0.5. The results are presented in Figs.~\ref{imbfulu1026}(a) and (b). Additionally, 
The imbalanced label distribution in the training data has a significant impact on the classification performance of the model~\cite{10155763,ZHOU2022108485}. We depicted the confusion matrices for the SST-2 and MR datasets under two imbalance levels in Figures~\ref{imbfulu1026}(c) and (d), in which the proportion of the negative class in demonstrations is set to 0.1 and 0.2. These results manifest that IDAICL significantly enhances the performance of the underrepresented classes in comparison to Vanilla ICL, thus proving its capability to address the class imbalance in demonstrations.

% we assessed IDAICL's ability to address class imbalance on two five-class classification datasets, namely SST-5 and Amazon, where class proportions were set at 0.05, 0.05, 0.1, 0.3, and 0.5. The same conclusion can be drawn from these experiments, highlighting that IDAICL significantly improves the accuracy of underrepresented classes.

% We presented the prediction probabilities for all classes achieved by Vanilla ICL and IDAICL on the GPT-2 model with 1.5B parameters within the Amazon and CB datasets. The results are demonstrated in Figs.\ref{confu}, highlighting the substantial enhancement in prediction accuracy accomplished through IDAICL, particularly for instances belonging to the minority classes. Consequently, IDAICL effectively mitigates class imbalance, thereby improving the fairness of models across different classes.

\section{Varying Demonstration Permutations}
Research has substantiated that the performance of ICL is sensitive to the permutation of demonstrations~\cite{lu-etal-2022-fantastically,zhao2021calibrate}. We assessed the performance of IDAICL under varying demonstration permutations. Specifically, we selected ten different sets of twelve training examples from the SST-2 datasets. For each set of examples, we shuffled the order ten times and calculated the accuracy for each permutation. 
% explore all 24 possible permutations of sample orders. 
The findings are depicted in Figure~\ref{shunxu}, indicating that IDAICL exhibits relatively stable performance across different demonstrations and permutations, while Vanilla ICL demonstrates high variance.

\begin{table*}[t]
% \vspace{-0.2in}
\centering
\small
\resizebox{1\linewidth}{!}{
\begin{tabular}{l|l|c}
\toprule[1.5pt]
\rowcolor{gray!30}
Format ID & Prompt & Label names \\
\midrule\midrule
   1       &   \makecell[l{p{11.5cm}}]{Review: This movie is amazing!\\
Answer: Positive\\
Review: Horrific movie, don’t see it.\\
Answer:}     &      Positive / Negative       \\ \hline
  2        &  \makecell[l{p{11.5cm}}]{Review: This movie is amazing!\\
Answer: good\\
Review: Horrific movie, don’t see it.\\
Answer:}      & good / bad   \\ \hline         3        &  \makecell[l{p{11.5cm}}]{My review for last night’s film: This movie is amazing! The critics agreed that this movie was good\\
My review for last night’s film: Horrific movie, don’t see it. The critics agreed that this movie was}      & good / bad   \\  \hline
   4        &  \makecell[l{p{11.5cm}}]{Here is what our critics think for this month’s films.\\
One of our critics wrote "This movie is amazing!". Her sentiment towards the film was positive.\\
One of our critics wrote "Horrific movie, don’t see it". Her sentiment towards the film was}      & positive / negative   \\ \hline 
   5        &  \makecell[l{p{11.5cm}}]{ Critical reception [ edit ]\\
In a contemporary review, Roger Ebert wrote "This movie is amazing!". Entertainment Weekly agreed, and
the overall critical reception of the film was good.\\
In a contemporary review, Roger Ebert wrote "Horrific movie, don’t see it". Entertainment Weekly agreed, and
the overall critical reception of the film was}      & good / bad   \\ \hline 
   6        &  \makecell[l{p{11.5cm}}]{Review: This movie is amazing!\\
Positive Review? Yes\\
Review: Horrific movie, don’t see it.\\
Positive Review?}      & Yes / No   \\ \hline 
   7        &  \makecell[l{p{11.5cm}}]{Review: This movie is amazing!\\
Question: Is the sentiment of the above review Positive or Negative?\\
Answer: Positive\\
Review: Horrific movie, don’t see it.\\
Question: Is the sentiment of the above review Positive or Negative?\\
Answer:}      & Positive / Negative  \\ \hline 
   8        &  \makecell[l{p{11.5cm}}]{ Review: This movie is amazing!\\
Question: Did the author think that the movie was good or bad?\\
Answer: good\\
Review: Horrific movie, don’t see it.\\
Question: Did the author think that the movie was good or bad?\\
Answer:}      &good / bad \\ \hline 
   9        &  \makecell[l{p{11.5cm}}]{Question: Did the author of the following tweet think that the movie was good or bad?\\
Tweet: This movie is amazing!\\
Answer: good\\
Question: Did the author of the following tweet think that the movie was good or bad?\\
Tweet: Horrific movie, don’t see it\\
Answer:}      &good / bad \\ \hline 
   10        &  \makecell[l{p{11.5cm}}]{This movie is amazing! My overall feeling was that the movie was good\\
Horrific movie, don’t see it. My overall feeling was that the movie was}      &good / bad \\ \hline 
   11        &  \makecell[l{p{11.5cm}}]{This movie is amazing! I liked the movie.\\
Horrific movie, don’t see it. I}      &liked / hated \\ \hline 
   12        &  \makecell[l{p{11.5cm}}]{ This movie is amazing! My friend asked me if I would give the movie 0 or 5 stars, I said 5\\
Horrific movie, don’t see it. My friend asked me if I would give the movie 0 or 5 stars, I said}      &0 / 5 \\ \hline 
   13        &  \makecell[l{p{11.5cm}}]{ Input: This movie is amazing!\\
Sentiment: Positive\\
Input: Horrific movie, don’t see it.\\
Sentiment:}      &Positive / Negative \\ \hline 
   14        &  \makecell[l{p{11.5cm}}]{Review: This movie is amazing!\\
Positive: True\\
Review: Horrific movie, don’t see it.\\
Positive:}      &True / False \\ \hline 
   15        &  \makecell[l{p{11.5cm}}]{ Review: This movie is amazing!\\
Stars: 5\\
Review: Horrific movie, don’t see it.\\
Stars:}      &5 / 0 \\ \bottomrule[1.5pt]
\end{tabular}}
\caption{The templates employed for examining the influence of formats on the SST-2 dataset, following those outlined by Zhao et al.~\shortcite{zhao2021calibrate}. An example from the training data is used for illustration.}
\label{formulate}
\end{table*}

\end{document}